\documentclass{article}

\PassOptionsToPackage{numbers}{natbib}

\usepackage[pagebackref=true]{hyperref} 
\renewcommand{\backref}[1]{} 
\renewcommand{\backrefalt}[4]{%
  \ifcase #1 
    (Not cited.)%
  \or 
    (Cited on page #4.)%
  \else 
    (Cited on pages #4.)%
  \fi%
}

\usepackage[preprint]{neurips_2026}
\usepackage[most]{tcolorbox}
\tcbuselibrary{skins}
\usepackage[utf8]{inputenc}
\usepackage[T1]{fontenc}
\usepackage{hyperref}
\usepackage{url}
\usepackage{booktabs}
\usepackage{amsfonts}
\usepackage{amsmath}
\usepackage{nicefrac}
\usepackage{microtype}
\usepackage{makecell}
\usepackage{xcolor}
\usepackage{graphicx}
\usepackage{multirow}
\usepackage{subcaption}
\usepackage{wrapfig}
\usepackage[table]{xcolor}
\usepackage{tcolorbox}

\hypersetup{colorlinks=true, linkcolor=NatGreen, citecolor=NatGreen, urlcolor=ditchblue}

\definecolor{bestcell}{RGB}{224,241,229}   
\newcommand{\ours}{\textsc{GeoPhys}}
\newcommand{\vect}[1]{\mathbf{#1}}

\definecolor{curvgreen}{HTML}{2CA02C}
\definecolor{natorange}{HTML}{FF7F0E}
\definecolor{accelblue}{HTML}{1F77B4}
\definecolor{predpurple}{HTML}{9467BD}
\definecolor{plausgreen}{HTML}{2CA02C} 
\definecolor{violred}{HTML}{D62728}

\usetikzlibrary{calc, arrows.meta, positioning, fit, backgrounds, decorations.pathreplacing}

\providecommand{\vect}[1]{\mathbf{#1}}

\definecolor{ditchblue}{HTML}{2E5C8A}     
\definecolor{ditchteal}{HTML}{0F8A7E}     
\definecolor{cornorange}{HTML}{C25A12}    
\definecolor{voneplum}{HTML}{6E3B8A}      
\definecolor{frostblue}{HTML}{4A90C9}     
\definecolor{inkgrey}{HTML}{2A2A2A}
\definecolor{softgrey}{HTML}{8C8C8C}
\definecolor{paleline}{HTML}{C8C8C8}

\newcommand{\snowflake}[1][0.10cm]{%
  \tikz[baseline=-0.55ex]{
    \foreach \a in {0,60,120,180,240,300}{
      \draw[frostblue, line width=0.5pt, line cap=round] (0,0) -- (\a:#1);
      \draw[frostblue, line width=0.4pt, line cap=round]
        (\a:0.55*#1) -- ++(\a-30:0.30*#1);
      \draw[frostblue, line width=0.4pt, line cap=round]
        (\a:0.55*#1) -- ++(\a+30:0.30*#1);
    }
    \fill[frostblue] (0,0) circle (0.04*#1);
  }%
}

\tikzset{
  framenode/.style={inner sep=0pt, outer sep=0pt, draw=paleline, line width=0.4pt},
  rowsub/.style={font=\tiny\itshape, text=softgrey},
  rowdim/.style={font=\tiny, text=softgrey},
  coltitle/.style={font=\scriptsize, text=black, anchor=center},
  archblock/.style={draw=#1, fill=#1!8, line width=0.45pt, rounded corners=0.4pt,
                    minimum width=0.32cm, minimum height=0.16cm, inner sep=0pt},
  archhilite/.style={draw=#1, fill=#1, line width=0.6pt, rounded corners=0.4pt,
                     minimum width=0.32cm, minimum height=0.16cm, inner sep=0pt},
  stage/.style={draw=#1, fill=#1!10, line width=0.55pt, rounded corners=1.5pt,
                minimum width=0.55cm, minimum height=0.42cm, inner sep=0pt,
                font=\tiny\bfseries, text=#1!50!black},
  stagehilite/.style={draw=#1, fill=#1!85, line width=0.7pt, rounded corners=1.5pt,
                      minimum width=0.55cm, minimum height=0.42cm, inner sep=0pt,
                      font=\tiny\bfseries, text=white},
  stagearr/.style={-{Stealth[length=1.4mm, width=1.0mm]}, line width=0.5pt, draw=softgrey,
                   shorten <=1pt, shorten >=1pt},
  reclooparr/.style={-{Stealth[length=1.0mm, width=0.8mm]}, line width=0.45pt, draw=#1!70,
                     shorten <=0.5pt, shorten >=0.5pt},
  fanoutarr/.style={-{Stealth[length=1.4mm, width=1.0mm]}, line width=0.7pt, draw=softgrey,
                    shorten <=2pt, shorten >=2pt},
  trajdot/.style={circle, fill=#1, draw=white, line width=0.3pt, inner sep=0.6pt},
  trajline/.style={#1, line width=1.2pt, line cap=round},
}

\newcommand{\vitstack}[2]{%
  \begin{tikzpicture}[baseline=(stack.center)]
    \def\bw{0.32}\def\bh{0.20}\def\gap{0.018}
    \foreach \i in {1,...,24}{
      \pgfmathsetmacro{\xx}{(\i-1)*(\bh+\gap)}
      \ifnum\i=#2
        \node[archhilite=#1, minimum width=\bh cm, minimum height=\bw cm]
          at (\xx, 0) (b\i) {};
      \else
        \node[archblock=#1, minimum width=\bh cm, minimum height=\bw cm]
          at (\xx, 0) (b\i) {};
      \fi
    }
    \node[fit=(b1)(b24), inner sep=0pt] (stack) {};
    \draw[stagearr, draw=#1, line width=0.6pt]
      ($(b#2.north)+(0,0.04)$) -- ++(0,0.18)
      node[anchor=south, font=\tiny\bfseries, text=#1, inner sep=0.5pt] {$\ell{=}#2$};
    \node[font=\tiny, text=softgrey, anchor=north, inner sep=1pt] at (b1.south) {$1$};
    \node[font=\tiny, text=softgrey, anchor=north, inner sep=1pt] at (b24.south) {$24$};
  \end{tikzpicture}%
}

\newcommand{\cornetschem}[2]{%
  \begin{tikzpicture}[baseline=(net.center)]
    \foreach \name/\xx in {V1/0, V2/0.80, V4/1.60, IT/2.40}{
      \ifx\name#2
        \node[stagehilite=#1] (\name) at (\xx, 0) {\name};
      \else
        \node[stage=#1] (\name) at (\xx, 0) {\name};
      \fi
    }
    \draw[stagearr] (V1.east) -- (V2.west);
    \draw[stagearr] (V2.east) -- (V4.west);
    \draw[stagearr] (V4.east) -- (IT.west);
    \foreach \name in {V1, V2, V4, IT}{
      \draw[reclooparr=#1] ($(\name.north)+(-0.10,0.02)$)
        .. controls +(-0.06,0.20) and +(0.06,0.20) ..
        ($(\name.north)+(0.10,0.02)$);
    }
    \draw[stagearr, draw=#1, line width=0.6pt]
      ($(IT.north)+(0,0.20)$) -- ++(0,0.20)
      node[anchor=south, font=\tiny\bfseries, text=#1, inner sep=0.5pt] {readout};
    \node[fit=(V1)(IT), inner sep=0pt] (net) {};
  \end{tikzpicture}%
}

\newcommand{\vonenetschem}[2]{%
  \begin{tikzpicture}[baseline=(net.center)]
    \node[draw=#1, fill=#1!85, line width=0.7pt, rounded corners=1.5pt,
          minimum width=0.85cm, minimum height=0.55cm, inner sep=0pt] (vone) at (0, 0) {};
    \begin{scope}
      \clip ($(vone.south west)+(0.04,0.04)$) rectangle ($(vone.north east)+(-0.04,-0.04)$);
      \foreach \a/\dx/\dy in {0/-0.24/0.10, 45/0.0/0.10, 90/0.24/0.10,
                              0/-0.24/-0.10, 45/0.0/-0.10, 90/0.24/-0.10}{
        \draw[white, line width=0.5pt, line cap=round]
          ($(vone.center)+(\dx,\dy)$) ++(\a:0.075) -- ++(\a+180:0.15);
      }
    \end{scope}
    \foreach \name/\xx in {L1/1.10, L2/1.65, L3/2.20, L4/2.75}{
      \node[stage=#1, minimum width=0.40cm] (\name) at (\xx, 0) {\name};
    }
    \draw[stagearr] (vone.east) -- (L1.west);
    \draw[stagearr] (L1.east) -- (L2.west);
    \draw[stagearr] (L2.east) -- (L3.west);
    \draw[stagearr] (L3.east) -- (L4.west);
    \draw[stagearr, draw=#1, line width=0.6pt]
      ($(vone.north)+(0,0.04)$) -- ++(0,0.20)
      node[anchor=south, font=\tiny\bfseries, text=#1, inner sep=0.5pt] {readout};
    \node[fit=(vone)(L4), inner sep=0pt] (net) {};
  \end{tikzpicture}%
}

\usetikzlibrary{calc, arrows.meta, positioning, fit, backgrounds, decorations.pathreplacing}

\providecommand{\vect}[1]{\mathbf{#1}}

\definecolor{ditchblue}{HTML}{2E5C8A}     
\definecolor{ditchteal}{HTML}{0F8A7E}     
\definecolor{cornorange}{HTML}{C25A12}    
\definecolor{voneplum}{HTML}{6E3B8A}      
\definecolor{frostblue}{HTML}{4A90C9}     
\definecolor{inkgrey}{HTML}{2A2A2A}
\definecolor{softgrey}{HTML}{8C8C8C}
\definecolor{paleline}{HTML}{C8C8C8}

\tikzset{
  framenode/.style={inner sep=0pt, outer sep=0pt, draw=paleline, line width=0.4pt},
  rowsub/.style={font=\tiny\itshape, text=softgrey},
  rowdim/.style={font=\tiny, text=softgrey},
  coltitle/.style={font=\scriptsize, text=black, anchor=center},
  archblock/.style={draw=#1, fill=#1!8, line width=0.45pt, rounded corners=0.4pt,
                    minimum width=0.32cm, minimum height=0.16cm, inner sep=0pt},
  archhilite/.style={draw=#1, fill=#1, line width=0.6pt, rounded corners=0.4pt,
                     minimum width=0.32cm, minimum height=0.16cm, inner sep=0pt},
  stage/.style={draw=#1, fill=#1!10, line width=0.55pt, rounded corners=1.5pt,
                minimum width=0.55cm, minimum height=0.42cm, inner sep=0pt,
                font=\tiny\bfseries, text=#1!50!black},
  stagehilite/.style={draw=#1, fill=#1!85, line width=0.7pt, rounded corners=1.5pt,
                      minimum width=0.55cm, minimum height=0.42cm, inner sep=0pt,
                      font=\tiny\bfseries, text=white},
  stagearr/.style={-{Stealth[length=1.4mm, width=1.0mm]}, line width=0.5pt, draw=softgrey,
                   shorten <=1pt, shorten >=1pt},
  reclooparr/.style={-{Stealth[length=1.0mm, width=0.8mm]}, line width=0.45pt, draw=#1!70,
                     shorten <=0.5pt, shorten >=0.5pt},
  fanoutarr/.style={-{Stealth[length=1.4mm, width=1.0mm]}, line width=0.7pt, draw=softgrey,
                    shorten <=2pt, shorten >=2pt},
  trajdot/.style={circle, fill=#1, draw=white, line width=0.3pt, inner sep=0.6pt},
  trajline/.style={#1, line width=1.2pt, line cap=round},
}

\newcommand{\trajA}[1]{%
  \begin{tikzpicture}[baseline=(box.center)]
    \coordinate (p1) at (0.00, 0.04); \coordinate (p2) at (0.28, 0.13);
    \coordinate (p3) at (0.58, 0.21); \coordinate (p4) at (0.90, 0.24);
    \coordinate (p5) at (1.22, 0.21); \coordinate (p6) at (1.52, 0.13);
    \coordinate (p7) at (1.80, 0.04);
    \draw[trajline=#1] plot[smooth, tension=0.55]
      coordinates {(p1)(p2)(p3)(p4)(p5)(p6)(p7)};
    \foreach \p in {p1,p2,p3,p4,p5,p6,p7}{ \node[trajdot=#1] at (\p) {}; }
    \node[font=\tiny, text=#1!60!black, anchor=south east, inner sep=0pt]
      at ($(p1)+(0.02,0.06)$) {$\bar{\vect{z}}_1$};
    \node[font=\tiny, text=#1!60!black, anchor=south west, inner sep=1pt]
      at ($(p7)+(0.02,0.05)$) {$\bar{\vect{z}}_T$};
    \node[fit=(p1)(p7), inner sep=2pt] (box) {};
  \end{tikzpicture}%
}

\newcommand{\trajB}[1]{%
  \begin{tikzpicture}[baseline=(box.center)]
    \coordinate (p1) at (0.00, 0.04); \coordinate (p2) at (0.30, 0.07);
    \coordinate (p3) at (0.60, 0.10); \coordinate (p4) at (0.90, 0.13);
    \coordinate (p5) at (1.20, 0.15); \coordinate (p6) at (1.50, 0.16);
    \coordinate (p7) at (1.80, 0.16);
    \draw[trajline=#1] plot[smooth, tension=0.55]
      coordinates {(p1)(p2)(p3)(p4)(p5)(p6)(p7)};
    \foreach \p in {p1,p2,p3,p4,p5,p6,p7}{ \node[trajdot=#1] at (\p) {}; }
    \node[font=\tiny, text=#1!60!black, anchor=south east, inner sep=0pt]
      at ($(p1)+(0.02,0.06)$) {$\bar{\vect{z}}_1$};
    \node[font=\tiny, text=#1!60!black, anchor=south west, inner sep=1pt]
      at ($(p7)+(0.02,0.05)$) {$\bar{\vect{z}}_T$};
    \node[fit=(p1)(p7), inner sep=2pt] (box) {};
  \end{tikzpicture}%
}

\newcommand{\trajC}[1]{%
  \begin{tikzpicture}[baseline=(box.center)]
    \coordinate (p1) at (0.00, 0.05); \coordinate (p2) at (0.22, 0.20);
    \coordinate (p3) at (0.50, 0.28); \coordinate (p4) at (0.85, 0.10);
    \coordinate (p5) at (1.15, 0.04); \coordinate (p6) at (1.45, 0.22);
    \coordinate (p7) at (1.80, 0.07);
    \draw[trajline=#1] plot[smooth, tension=0.55]
      coordinates {(p1)(p2)(p3)(p4)(p5)(p6)(p7)};
    \foreach \p in {p1,p2,p3,p4,p5,p6,p7}{ \node[trajdot=#1] at (\p) {}; }
    \node[font=\tiny, text=#1!60!black, anchor=south east, inner sep=0pt]
      at ($(p1)+(0.02,0.06)$) {$\bar{\vect{z}}_1$};
    \node[font=\tiny, text=#1!60!black, anchor=south west, inner sep=1pt]
      at ($(p7)+(0.02,0.05)$) {$\bar{\vect{z}}_T$};
    \node[fit=(p1)(p7), inner sep=2pt] (box) {};
  \end{tikzpicture}%
}

\newcommand{\trajD}[1]{%
  \begin{tikzpicture}[baseline=(box.center)]
    \coordinate (p1) at (0.00, 0.10); \coordinate (p2) at (0.20, 0.24);
    \coordinate (p3) at (0.42, 0.06); \coordinate (p4) at (0.66, 0.22);
    \coordinate (p5) at (0.90, 0.08); \coordinate (p6) at (1.16, 0.24);
    \coordinate (p7) at (1.42, 0.10); \coordinate (p8) at (1.65, 0.20);
    \coordinate (p9) at (1.80, 0.13);
    \draw[trajline=#1] plot[smooth, tension=0.55]
      coordinates {(p1)(p2)(p3)(p4)(p5)(p6)(p7)(p8)(p9)};
    \foreach \p in {p1,p2,p3,p4,p5,p6,p7,p8,p9}{ \node[trajdot=#1] at (\p) {}; }
    \node[font=\tiny, text=#1!60!black, anchor=south east, inner sep=0pt]
      at ($(p1)+(0.02,0.06)$) {$\bar{\vect{z}}_1$};
    \node[font=\tiny, text=#1!60!black, anchor=south west, inner sep=1pt]
      at ($(p9)+(0.02,0.05)$) {$\bar{\vect{z}}_T$};
    \node[fit=(p1)(p9), inner sep=2pt] (box) {};
  \end{tikzpicture}%
}

   \usepackage[table]{xcolor}
   \usepackage[framemethod=TikZ]{mdframed}
\usepackage{xcolor}
\usepackage{mdframed}
\definecolor{NatGreen}{RGB}{46,125,50}     
   \definecolor{bestcell}{RGB}{224,241,229}   
   \newcommand{\best}[1]{\cellcolor{bestcell}\textbf{#1}}

\newcounter{takeawaycounter}

\tcbset{
  takeawaybox/.style={
    enhanced,
    colback=white,                         
    colframe=black!85,                     
    boxrule=1pt,
    borderline={0.4pt}{2.5pt}{NatGreen},   
    drop fuzzy shadow=black!10,
    arc=4pt,
    left=22pt, right=12pt, top=10pt, bottom=10pt,
    boxsep=0pt,
    width=\linewidth,
    coltext=black,
    overlay={
      \node[text=NatGreen, font=\normalsize]
        at ([xshift=12pt, yshift=-13pt]frame.north west)
        {\faArrowRight};
    }
  }
}

\newenvironment{takeaway}{%
  \stepcounter{takeawaycounter}%
  \begin{tcolorbox}[takeawaybox]%
    \textbf{\textcolor{NatGreen}{Takeaway~\thetakeawaycounter.}}\hspace{0.4em}%
}{%
  \end{tcolorbox}%
}

\usepackage{tikz}
\usetikzlibrary{angles, quotes, calc, arrows.meta, decorations.pathreplacing, decorations.markings}
\usepackage{caption}    
\usepackage{fontawesome5}
\definecolor{NatGreen}{RGB}{46,125,50}

\newcounter{hypothesis}

\tcbset{
  hypobox/.style={
    enhanced,
    colback=NatGreen!8,
    colframe=NatGreen,
    boxrule=1pt,                  
    arc=3pt,                      
    left=22pt,                    
    right=8pt, top=5pt, bottom=5pt,  
    boxsep=0pt,
    width=\linewidth,
    overlay={
      \node[text=NatGreen, font=\normalsize] at ([xshift=12pt, yshift=-10pt]frame.north west) {\faLightbulb};
    }
  }
}
\definecolor{HypoBlue}{RGB}{25,118,210}

\definecolor{plausgreen}{RGB}{ 70,160, 90}    
\definecolor{violred}{RGB}{200, 70, 70}       
\definecolor{curvgreen}{RGB}{205,230,207}     
\definecolor{natorange}{RGB}{240,170, 90}     
\definecolor{accelblue}{RGB}{120,170,220}     
\definecolor{predpurple}{RGB}{180,150,210}    

\tcbset{
  hypobox/.style={
    enhanced,
    colback=white,                         
    colframe=black!85,                     
    boxrule=1pt,
    borderline={0.4pt}{2.5pt}{NatGreen},   
    drop fuzzy shadow=black!10,
    arc=4pt,
    left=12pt, right=12pt, top=10pt, bottom=10pt,
    boxsep=0pt,
    width=\linewidth,
    coltext=black
  }
}

\title{\ours{}: The Geometry of Physical Plausibility}

\author{%
  \textbf{Christian Internò}$^{1}$\thanks{Correspondence to: \texttt{christianintern17@gmail.com}\\
\faGlobe~\textbf{Project page:} \url{https://christianinterno.github.io/GeoPhys/}\\
\faGithub~\textbf{Code:} \url{https://github.com/ChristianInterno/GeoPhys}} \quad
  \textbf{Alexander Pondaven}$^{2}$ \quad
  \textbf{Habon Issa}$^{3}$ \quad
  \textbf{Fabio Pizzati}$^{4}$\\
  \textbf{Francesco Pinto}$^{5}$ \quad
  \textbf{Markus Olhofer}$^{6}$ \quad
  \textbf{Ivan Laptev}$^{4}$ \quad
  \textbf{Philip Torr}$^{2}$\\
  \textbf{Eero P.~Simoncelli}$^{7,8}$ \quad
  \textbf{Barbara Hammer}$^{1}$ \quad
  \textbf{David Klindt}$^{3}$\\[10pt]
  $^1$ Bielefeld University \quad
  $^2$ University of Oxford \quad
  $^3$ Cold Spring Harbor Laboratory\\
  $^4$ MBZUAI \quad 
  $^5$ Independent \quad
  $^6$ Honda Research Institute EU \\
  $^7$ New York University \quad
  $^8$ Flatiron Institute, Simons Foundation \\[5pt]
}
\usepackage{capt-of}
\usepackage{fancyhdr}
\hypersetup{colorlinks=true, linkcolor=NatGreen, citecolor=NatGreen, urlcolor=NatGreen}

\begin{document}

\maketitle

\begin{abstract}
While humans can identify physically implausible events within milliseconds, machine
learning approaches addressing the same problem are extremely slow and expensive. They
either rely on external multimodal-LLM judges or require ad-hoc modifications to the
training procedure. In this work, we argue that indicators of physical plausibility are
implicitly captured by five geometric properties of the per-frame embeddings produced by
frozen image encoders. In aggregate, we call them \ours{}. First, we show that these
signals correlate with human EEG responses to two forms of object-permanence violations. Second, \ours{} robustly discriminates physically
implausible videos from realistic ones, achieving state-of-the-art physics-violation
detection: $98.3\%$ on LikePhys~\cite{yuan2025likephys} and $93.3\%$ on IntPhys2~\cite{bordes2025intphys2}, whereas V-JEPA~2, GPT-4o,
Gemini, and twelve modern video diffusion models perform near chance. Third, used as a best-of-$N$ verifier for physical alignment during video generation, \ours{} lifts MAGI-1~24B from 50.01\% to 64.50\% on
PhysicsIQ~\cite{motamed2025physicsiq} at 1.5$\times$ lower wall-clock and 4.65$\times$
lower memory than the V-JEPA~2 world-model verifier. Ultimately, \ours{} demonstrates that physical plausibility in videos can be assessed by leveraging the emergent geometric properties of temporal features extracted from image encoders.
\end{abstract}

\section{Introduction}
\label{sec:intro}


Humans readily detect physical violations directly from visual perception: infants register surprise on seeing impossible events~\cite{margoni2024violation,dunn2017investigating}, and adult brains register such events rapidly, without
deliberation~\cite{balaban2024electrophysiology,liu2024violations}. In machine learning, by contrast, detecting physical violations in video has been achieved using physics-targeted model training~\cite{garrido2025intuitivephysicsunderstandingemerges,wang2025physcorrdualrewarddpophysicsconstrained}, billion-parameter video-pretrained world models~\cite{yuan2026wmreward,assran2025vjepa2}, multi-modal language models prompted to act as judges~\cite{motamed2025travlrecipemakingvideolanguage}, neuro-symbolic pipelines that parse scenes and run physical
simulators~\cite{smith2019adept, battaglia2016interaction, wu2015galileo} or hand-crafted forensic checks for physical
consistency in generated imagery~\cite{farid2022lightinginconsistencypainttext,102629646,Barrington_2026_CVPR,farid2022perspectiveinconsistencypainttext}. All these approaches share the core assumption that signals for physical plausibility must be explicitly induced into, or reasoned about by a model. In this work, we challenge that assumption by asking a simpler question: 
\begin{quote}
    \centering
    \emph{Does a vision model trained on still images, with no video training or physics supervision, contain a geometric signature of physical plausibility?}
\end{quote}
A long line of work in machine learning and computational neuroscience suggests that good visual representations should make natural transformations geometrically simple, mapping nonlinear pixel-space motion onto approximately linear feature-space trajectories. This thread runs from temporal slowness and tangent-propagation~\cite{foldiak1991learning,wiskott2002slow,simard1991tangentprop} through Lie-group representations of transformation~\cite{rao1998learning} to learned linearization~\cite{goroshin2015linearize,henaff2016geodesics,niu2024learning}, and connects to the neuroscience finding that primate visual cortex straightens natural-video trajectories relative to pixel space~\cite{henaff2019straightening,henaff2021straightening,harrington2023exploring,zonneveld2025straightening}. Together, these
results motivate a concrete prediction about physical plausibility:

\begin{tcolorbox}[hypobox]
\refstepcounter{hypothesis}\label{box:hypothesis}
\textbf{\textcolor{NatGreen}{\faLightbulb\quad Hypothesis.}}
\\
\textit{
Vision models learn to linearize natural video dynamics, mapping physically plausible videos to smooth, locally predictable feature-space trajectories disrupted by physical violations.
}
\end{tcolorbox}

This link between geometry and physical plausibility is purely statistical.
We do not argue that frozen backbones represent physics or reason about
it~\citep{vafa2025foundationmodelfoundusing,liu2024violations,
interno2026observer}; we argue that the geometry of feature-space
trajectories is disrupted during physics-violations in existing
benchmarks~\cite{bordes2025intphys2,yuan2025likephys,liu2024violations}
reliably enough to be useful. We find that this prediction holds across four distinct backbones, none trained on video or physics:
self-supervised transformers (DINOv2~\cite{oquab2024dinov2} \& v3~\cite{simoni2025dinov3}), a recurrent ventral-stream CNN
(CORnet-S~\cite{kubilius2019cornet}), and a ResNet with a fixed Gabor
V1 front-end (VOneNet~\cite{vonenet2020}).

Building on our insights, we introduce \ours{}, a training-free score that captures the temporal evolution of feature trajectories. It computes a set of kinematic statistics on feature space across frames: speed and its variation, turning-angle curvature, acceleration, and the residual of a linear auto-regressive predictor. These combine into a single scalar physical plausibility score. We apply the same score across three settings, each testing a property a useful signal should have: \textbf{(i)}~It must correspond to a real phenomenon rather than an artifact of pretraining statistics. We test this by examining the score's temporal response to physical violations and its alignment with human neural responses. \textbf{(ii)}~It must discriminate plausible from violated video at scale. We test this via detection on standard physics-violation benchmarks. \textbf{(iii)}~It must transfer beyond passive measurement to active control. We test this via inference-time best-of-$N$ selection for physically plausible video generation.

\paragraph{Contributions:}
\textbf{(1)~Biologically grounded}. Lacking pretraining on physics objectives, geometric signals could flag invalid videos for reasons unrelated to physical violations per se. We thus validate per-frame \ours{} signals are directionally meaningful, aligning with known surprise-detection signatures of human EEG responses to matched stimuli~\cite{balaban2024electrophysiology,liu2024violations,kaiser2023eeg}. \textbf{(2)~Detection}. We
achieve SoTA physics-violation detection: $98.3\%$ on
LikePhys~\cite{yuan2025likephys} and $93.3\%$ on
IntPhys2~\cite{bordes2025intphys2}. Every individual backbone exceeds
all published baselines, including V-JEPA~2~\cite{assran2025vjepa2},
GPT-4o~\cite{openai2024gpt4technicalreport},
Gemini~\cite{geminiteam2025geminifamilyhighlycapable}, and twelve
video diffusion models. \textbf{(3)~Generation}. We offer the signal
as an inference-time verifier for test-time scaling. It closes more
of the gap to the oracle ceiling than any prior method on a matched
candidate pool. On PhysicsIQ~\cite{motamed2025physicsiq}, it lifts
MAGI-1~24B~\cite{magi1} from $50.01\%$ to $64.50\%$. This comes at
$1.5\times$ lower wall-clock and $4.65\times$ lower memory than the VJEPA-2 baseline WMReward~\cite{yuan2026wmreward}.



\section{Related work}
\label{sec:related}
\vspace{-0.4em}
\textbf{Physics evaluation in video models.}
Physical-plausibility benchmarks include PhysicsIQ~\cite{motamed2025physicsiq}
(real-world fidelity), LikePhys~\cite{yuan2025likephys} ($12$ dynamic scenarios),
and IntPhys2~\cite{bordes2025intphys2} (four core-knowledge properties), after
earlier Physion~\cite{bear2021physion}, IntPhys~\cite{riochet2020intphys}, and
GRASP~\cite{jassim2024grasp}. Three baseline families dominate: multimodal-LLM
judges~\cite{motamed2025travlrecipemakingvideolanguage,
openai2024gpt4technicalreport,geminiteam2025geminifamilyhighlycapable}, video
diffusion models scored by likelihood~\cite{wan2025,hunyuan2024,cogvideox2024},
and video-pretrained encoders~\cite{assran2025vjepa2,videomaev2}. All require
video-scale pretraining, billion-parameter inference, or both. \ours{} requires
neither: every backbone is a frozen image encoder, sub-billion parameters, no
video pretraining.

\textbf{Verifier models for video generation.}
Inference-time methods rerank a fixed generator's candidates with a
verifier~\cite{liu2025video}. WMReward~\cite{yuan2026wmreward}, the SoTA on
PhysicsIQ, uses V-JEPA~2~\cite{assran2025vjepa2} ($1.1$B parameters,
video-pretrained). Other verifiers use multimodal-LLM
judges~\cite{yang2025qwen3technicalreport}, video-feature
backbones~\cite{videomaev2}, learned video-quality
models~\cite{he2024videoscore,bansal2024videophy}, or 3D scene-geometry
consistency~\cite{yin2026vigorvideogeometryorientedreward}. All draw the signal from a video-pretrained,
language-scale, or 3D-estimation system. We use the opposite: a frozen image
encoder with no learned ranker. This places physics-aware generation in the
test-time scaling literature~\cite{cobbe2021training,lightman2023verify,
snell2024scaling,brown2024large,singhal2025a,oshima2026inferencetime,
Baraldi_2025_BMVC}, where verifier cost trades off against selection quality.

\textbf{Geometric analysis of feature trajectories.}
The straightening lineage~\cite{foldiak1991learning,wiskott2002slow,
rao1998learning,simard1991tangentprop,goroshin2015linearize,
henaff2021straightening,harrington2023exploring,zonneveld2025straightening,
niu2024learning} studies feature-trajectory geometry in biological and
artificial vision. Predictive coding offers a complementary
view~\cite{rao1999predictive,friston2010freeenergy}: perception minimises
feature-space prediction error, motivating \ours{}'s residual signal. For
AI-vs-natural-video detection, ReStraV~\cite{interno2026aigenerated} uses
DINOv2 curvature and Grab-3D~\cite{chen2025grab3ddetectingaigeneratedvideos}
uses 3D scene-geometry consistency. \ours{} is the first to apply
frozen-image-encoder trajectory geometry to physical plausibility itself,
across four backbones (ViTs~\cite{oquab2024dinov2} and V1-like
CNNs~\cite{vonenet2020}), five kinematic statistics, and three settings:
neural alignment~(\ref{sec:neuro}), physics-violation
benchmarks~(\ref{sec:exp_detection}), and best-of-$N$ selection at generation
time~(\ref{sec:exp_bon}).

\section{A \textsc{Geo}metrical signal of \textsc{Phy}sical plausibility}
\label{sec:method_signals}

\begin{figure}[htb]
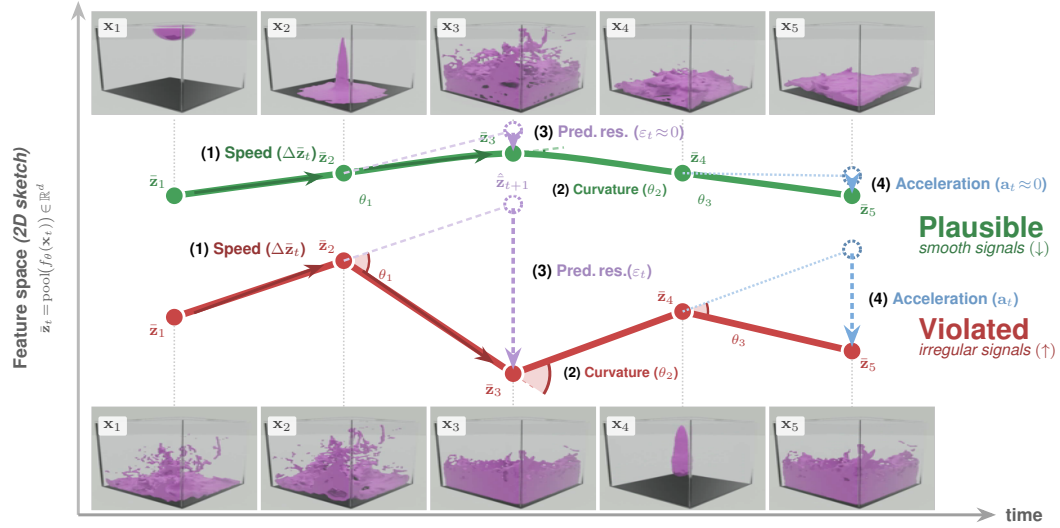

\centering

\resizebox{\linewidth}{!}{%
\begin{tikzpicture}[
    >=Stealth,
    plauspoint/.style={circle, fill=plausgreen, draw=white,
                       line width=1.5pt, inner sep=3.5pt},
    violpoint/.style={circle, fill=violred, draw=white,
                       line width=1.5pt, inner sep=3.5pt},
    ghostpoint/.style={circle, draw=predpurple, fill=white,
                       line width=1.5pt, inner sep=3.0pt, densely dotted},
    thumbframe/.style={inner sep=0pt, outer sep=0pt,
                       draw=black!25, line width=1.0pt},
    rowlabel/.style={font=\Large\bfseries\sffamily},
    rowsublabel/.style={font=\footnotesize\sffamily\itshape},
    midlabel/.style={font=\small\bfseries\sffamily},
    minilabel/.style={font=\scriptsize\bfseries\sffamily},   
    pointlabel/.style={font=\footnotesize\bfseries\sffamily}, 
    arr/.style={-Stealth, line width=1.5pt, shorten <=3pt, shorten >=3pt},
    cleanlabel/.style={fill=white, fill opacity=0.85, text opacity=1, inner sep=2.5pt, rounded corners=1.5pt}
]


\draw[-Stealth, gray!55, line width=1.5pt] (-0.7,-2.70) -- (-0.7, 6.40);
\draw[-Stealth, gray!75, line width=1.5pt] (-0.7,-2.70) -- (15.55,-2.70);
\node[midlabel, gray!75!black, anchor=west] at (15.60,-2.70) {time};

\node[midlabel, gray!75!black, anchor=south, rotate=90, align=center]
  at (-1.00, 1.85)
  {Feature space \textit{(2D sketch)}\\[1.5pt]
   {\scriptsize $\bar{\vect{z}}_t \!=\! \mathrm{pool}\!\bigl(f_\theta(\vect{x}_t)\bigr)\!\in\!\mathbb{R}^{d}$}};

\foreach \i/\x in {1/1.0, 2/4.0, 3/7.0, 4/10.0, 5/13.0}{
  \node[thumbframe] at (\x, 5.30)
    {\includegraphics[width=2.9cm,height=1.8cm]{figures/thumbs/plaus_f/plaus_f\i.png}};
  \node[font=\footnotesize\bfseries, anchor=north west, inner sep=2.5pt,
        fill=white, fill opacity=0.85, text opacity=1,
        rounded corners=1pt, text=black!75]
        at ($(\x, 5.20)+(-1.35, 0.9)$) {$\vect{x}_{\i}$};
}
\foreach \x/\gy in {1.0/2.95, 4.0/3.35, 7.0/3.70, 10.0/3.35, 13.0/2.95}{
  \draw[gray!40, densely dotted, line width=0.8pt] (\x, 4.30) -- (\x,\gy+0.12);
}

\coordinate (g1) at (1.0,  2.95);
\coordinate (g2) at (4.0,  3.35);
\coordinate (g3) at (7.0,  3.70);
\coordinate (g4) at (10.0, 3.35);
\coordinate (g5) at (13.0, 2.95);

\coordinate (gext2) at ($(g2)!-0.25!(g1)$);
\draw[plausgreen!50, densely dashed, line width=1.0pt] (g2) -- (gext2);
\pic[draw=plausgreen!85!black, fill=plausgreen!22, line width=1.0pt, angle radius=0.40cm]
   {angle = g3--g2--gext2};

\coordinate (gext3) at ($(g3)!-0.30!(g2)$);
\draw[plausgreen!55, densely dashed, line width=1.2pt] (g3) -- (gext3);
\pic[draw=plausgreen!85!black, fill=plausgreen!22, line width=1.2pt, angle radius=0.55cm]
   {angle = g4--g3--gext3};

\coordinate (gext4) at ($(g4)!-0.25!(g3)$);
\draw[plausgreen!50, densely dashed, line width=1.0pt] (g4) -- (gext4);
\pic[draw=plausgreen!85!black, fill=plausgreen!22, line width=1.0pt, angle radius=0.40cm]
   {angle = g5--g4--gext4};

\draw[plausgreen, line width=3.2pt, line cap=round]
  plot[smooth, tension=0.5] coordinates {(g1)(g2)(g3)(g4)(g5)};
\foreach \p in {g1,g2,g3,g4,g5}{ \node[plauspoint] at (\p) {}; }

\node[rowlabel, plausgreen!85!black, anchor=west] at (14.05, 2.4) {Plausible};
\node[rowsublabel, plausgreen!75!black, anchor=west] at (14.05, 2.0)
  {smooth signals $(\downarrow)$};

\node[pointlabel, plausgreen!85!black, anchor=south east, inner sep=1pt] at ($(g1)+(-0.10, 0.10)$) {$\bar{\vect{z}}_1$};
\node[pointlabel, plausgreen!85!black, anchor=south east, inner sep=1pt] at ($(g2)+(-0.10, 0.12)$) {$\bar{\vect{z}}_2$};
\node[pointlabel, plausgreen!85!black, anchor=south east, inner sep=1pt] at ($(g3)+(-0.25, 0.12)$) {$\bar{\vect{z}}_3$};
\node[pointlabel, plausgreen!85!black, anchor=south west, inner sep=1pt] at ($(g4)+(0.10, 0.12)$) {$\bar{\vect{z}}_4$};
\node[pointlabel, plausgreen!85!black, anchor=south west, inner sep=1pt] at ($(g5)+(0.10, -0.40)$) {$\bar{\vect{z}}_5$};

\draw[arr, plausgreen!75!black] ($(g1)+(0.25,0.04)$) -- ($(g2)+(-0.25,-0.04)$);
\node[pointlabel, plausgreen!75!black]
  at ($(g1)!0.5!(g2)+(0,0.55)$) {\textbf{\textcolor{black}{(1)} Speed} ($\Delta\bar{\vect{z}}_t$)};

\draw[arr, plausgreen!75!black] ($(g2)+(0.25,0.06)$) -- ($(g3)+(-0.25,0.0)$);

\node[minilabel, plausgreen!85!black, anchor=north] at ($(g2)+(0.40,-0.25)$) {$\theta_1$};
\node[minilabel, plausgreen!85!black, anchor=west, cleanlabel] at ($(g3)+(0.60, -0.65)$) {\textbf{\textcolor{black}{(2)} Curvature} ($\theta_2$)};
\node[minilabel, plausgreen!85!black, anchor=north] at ($(g4)+(0.40,-0.25)$) {$\theta_3$};

\coordinate (predg3) at (7.0, 4.10);
\draw[predpurple!50, densely dashed, line width=1.2pt] (g2) -- (predg3);
\node[ghostpoint] at (predg3) {};
\draw[-Stealth, predpurple, densely dashed, line width=1.8pt, shorten <=1pt, shorten >=1pt] (predg3) -- (g3);
\node[pointlabel, predpurple!90!black, anchor=south]
  at ($(predg3)+(1.7,-0.3)$) {\textbf{\textcolor{black}{(3)} Pred.\,res.} ($\varepsilon_t \!\approx\! 0$)};

\coordinate (predga) at (13.0, 3.30);
\draw[accelblue!55, densely dotted, line width=1.2pt] (g4) -- (predga);
\node[ghostpoint, draw=accelblue!80!black] at (predga) {};
\draw[-Stealth, accelblue, densely dashed, line width=1.8pt, shorten <=1pt, shorten >=1pt] (predga) -- (g5);
\node[pointlabel, accelblue!90!black, anchor=west, cleanlabel]
  at ($(predga)+(0.25,-0.15)$) {\textbf{\textcolor{black}{(4)} Acceleration} ($\vect{a}_t \!\approx\! 0$)};

\coordinate (r1) at (1.0,  0.80);
\coordinate (r2) at (4.0,  1.80);
\coordinate (r3) at (7.0, -0.20);
\coordinate (r4) at (10.0, 0.90);
\coordinate (r5) at (13.0, 0.20);
\coordinate (predr3) at ($(r2)+(r2)-(r1)$); 
\coordinate (preda)  at ($(r4)+(r4)-(r3)$); 

\pic[draw=violred!85!black, fill=violred!22, line width=1.2pt, angle radius=0.45cm]
   {angle = r3--r2--predr3};

\coordinate (rext3) at ($(r3)!-0.12!(r2)$);
\draw[violred!55, densely dashed, line width=1.0pt] (r3) -- (rext3);
\pic[draw=violred!85!black, fill=violred!22, line width=1.5pt, angle radius=0.65cm]
   {angle = rext3--r3--r4};

\pic[draw=violred!85!black, fill=violred!22, line width=1.2pt, angle radius=0.45cm]
   {angle = r5--r4--preda};

\draw[violred, line width=3.4pt, line cap=round, line join=miter, miter limit=8]
  (r1) -- (r2) -- (r3) -- (r4) -- (r5);
\foreach \p in {r1,r2,r3,r4,r5}{ \node[violpoint] at (\p) {}; }

\node[rowlabel, violred!90!black, anchor=west] at (14.05, 0.60) {Violated};
\node[rowsublabel, violred!80!black, anchor=west] at (14.05, 0.20)
  {irregular signals $(\uparrow)$};

\node[pointlabel, violred!90!black, anchor=north east, inner sep=1pt] at ($(r1)+(-0.10, -0.10)$) {$\bar{\vect{z}}_1$};
\node[pointlabel, violred!90!black, anchor=south east, inner sep=1pt] at ($(r2)+(-0.10, 0.10)$) {$\bar{\vect{z}}_2$};
\node[pointlabel, violred!90!black, anchor=north east, inner sep=1pt] at ($(r3)+(-0.15, -0.15)$) {$\bar{\vect{z}}_3$}; 
\node[pointlabel, violred!90!black, anchor=south east, inner sep=1pt] at ($(r4)+(-0.10, 0.10)$) {$\bar{\vect{z}}_4$};
\node[pointlabel, violred!90!black, anchor=north west, inner sep=1pt] at ($(r5)+(0.10, -0.10)$) {$\bar{\vect{z}}_5$};

\draw[arr, violred!75!black] ($(r1)+(0.25,0.04)$) -- ($(r2)+(-0.25,-0.04)$);
\node[pointlabel, violred!75!black]
  at ($(r1)!0.5!(r2)+(-0.20,0.70)$) {\textbf{\textcolor{black}{(1)} Speed} ($\Delta\bar{\vect{z}}_t$)};

\draw[arr, violred!75!black] ($(r2)+(0.25,-0.12)$) -- ($(r3)+(-0.25,0.12)$);

\draw[predpurple!50, densely dashed, line width=1.2pt] (r2) -- (predr3);
\node[ghostpoint] at (predr3) {};
\node[pointlabel, predpurple!90!black, anchor=south, cleanlabel]
  at ($(predr3)+(0.0, 0.15)$) {$\hat{\bar{\vect{z}}}_{t+1}$};
\draw[-Stealth, predpurple, densely dashed, line width=1.8pt, shorten <=2pt, shorten >=2pt] (predr3) -- (r3);
\node[pointlabel, predpurple!90!black, anchor=west, cleanlabel]
  at ($(predr3)!0.4!(r3)+(0.25,0.0)$) {\textbf{\textcolor{black}{(3)} Pred.\,res.}($\varepsilon_t$)};

\node[minilabel, violred!90!black, anchor=center] at ($(r2)+(0.75,-0.25)$) {$\theta_1$};
\node[minilabel, violred!90!black, anchor=west, cleanlabel] at ($(r3)+(0.80, 0.00)$) {\textbf{\textcolor{black}{(2)} Curvature}  ($\theta_2$)};
\node[minilabel, violred!90!black, anchor=center] at ($(r4)+(1,-0.5)$) {$\theta_3$};

\draw[accelblue!55, densely dotted, line width=1.2pt] (r4) -- (preda);
\node[ghostpoint, draw=accelblue!80!black] at (preda) {};
\draw[-Stealth, accelblue, densely dashed, line width=1.8pt, shorten <=2pt, shorten >=2pt] (preda) -- (r5);
\node[pointlabel, accelblue!90!black, anchor=west, cleanlabel]
  at ($(preda)!0.5!(r5)+(0.20,0.0)$) {\textbf{\textcolor{black}{(4)} Acceleration} ($\vect{a}_t$)};

\foreach \x/\ry in {1.0/0.80, 4.0/1.80, 7.0/-0.20, 10.0/0.90, 13.0/0.20}{
  \draw[gray!40, densely dotted, line width=0.8pt]
    (\x, \ry-0.15) -- (\x, -0.80);
}
\foreach \i/\x in {1/1.0, 2/4.0, 3/7.0, 4/10.0, 5/13.0}{
  \node[thumbframe] at (\x, -1.70)
    {\includegraphics[width=2.9cm,height=1.8cm]{figures/thumbs/viol_f/viol_f\i.png}};
  \node[font=\footnotesize\bfseries, anchor=north west, inner sep=2.5pt,
        fill=white, fill opacity=0.85, text opacity=1,
        rounded corners=1pt, text=black!75]
        at ($(\x, -1.70)+(-1.35, 0.75)$) {$\vect{x}_{\i}$};
}

\end{tikzpicture}%
}%
\caption{\textbf{\ours{} signals of plausible vs.\ violated dynamics
in frozen feature space.}
Paired counterfactuals from
LikePhys~\cite{yuan2025likephys}, rendered in
Blender~\cite{Blender2018}. A frozen backbone maps each
frame $\vect{x}_t$ to a pooled feature $\bar{\vect{z}}_t$,
yielding a trajectory in representation space (sketched in 2D).
\textcolor{plausgreen!85!black}{\textbf{Plausible}} videos produce
smooth trajectories; \textcolor{violred!90!black}{\textbf{Violated}} (no momentum conservation) ones show irregular. \textbf{(1)}~speed, \textbf{(2)}~curvature,
\textbf{(3)}~prediction residual, and \textbf{(4)}~acceleration.}
\label{fig:curvature_illustration}
\end{figure}

\ours{} is applied to synthetic and AI generated videos~\cite{melnik2024videodiffusionsurvey}, which systematically violate the physical laws inherent in real data~\cite{motamed2025physicsiq, bordes2025intphys2}. Given a $T$-frame video $\vect{V}=\{\vect{x}_1,\ldots,\vect{x}_T\}$ and a frozen visual backbone $f_\theta$, each frame yields spatial embeddings $Z_t = f_\theta(\vect{x}_t) = \{\vect{z}_{t,n}\}_{n=1}^{N} \in \mathbb{R}^{N\times d}$, containing $N$ tokens of dimension $d$. Spatial average-pooling extracts a single feature vector per frame, $\bar{\vect{z}}_t = \tfrac{1}{N}\sum_{n=1}^{N} \vect{z}_{t,n} \in \mathbb{R}^{d}$, defining the video as a discrete feature-space trajectory: $\Gamma_\theta(\vect{V}) = (\bar{\vect{z}}_1, \ldots, \bar{\vect{z}}_T)$. We evaluate temporal irregularities via finite differences on $\Gamma_\theta(\vect{V})$, using kinematic terms (``velocity'', ``speed'', ``acceleration'') strictly as geometric analogies in representation space, not as physical motion in pixels or world coordinates.

\subsection{GeoPhys signals}

\paragraph{First-order motion: speed.}
The trajectory's first-order kinematics are defined by the frame-to-frame feature displacement, $\vect{v}_t = \bar{\vect{z}}_{t+1} - \bar{\vect{z}}_t$ $(t=1,\ldots,T-1)$. \textit{Its magnitude, $s_t = \|\vect{v}_t\|_2$, measures the representational shift between consecutive frames (\textbf{(1)}, $\Delta\bar{\vect{z}}_t$ in Fig.~\ref{fig:curvature_illustration}).} Because stable video dynamics preclude abrupt step-size fluctuations, we summarize speed instability via its temporal standard deviation: $\phi_{\mathrm{speed}}(\vect{V}) = \operatorname{std}\bigl(\{s_t\}_{t=1}^{T-1}\bigr)$. A large $\phi_{\mathrm{speed}}$ flags erratic representational changes, such as objects jumping, disappearing, or moving inconsistently.

\paragraph{Curvature and angle consistency.}
\vspace{-0.3em}
Constant speed does not guarantee a straight trajectory: representations
can bend sharply between frames, marking locally nonlinear
transformations. Curvature is the central diagnostic in biological
vision~\cite{henaff2021straightening}: humans and macaque V1 populations
perceptually straighten natural videos while curving unnatural ones.
In artificial networks, straightening requires specific conditions (SSL
objectives~\cite{zonneveld2025straightening}, adversarial
robustness~\cite{garrido2025intuitivephysicsunderstandingemerges}) and is
absent from standard classifiers~\cite{henaff2019straightening}.
\textit{We measure local curvature by the turning angle between consecutive
displacements (\textbf{(2)}, $\theta_t$ in Fig.~\ref{fig:curvature_illustration}):}$
\theta_t = \arccos\!
    \left(
    \frac{\langle \vect{v}_t,\;
                  \vect{v}_{t+1} \rangle}
         {\|\vect{v}_t\|_2 \,
          \|\vect{v}_{t+1}\|_2}
    \right),
    \quad t = 1,\ldots,T-2$.
Locally linear transformations yield small $\theta_t$. Physical
violations break this linearity: a backbone expects natural dynamics
(balls bouncing off walls), so an implausible ``wall-pass'' frame
diverges sharply from the smooth pre-impact trajectory, bending
$\theta_t$. We capture this with two statistics: the mean turning angle
$\phi_{\mathrm{curv}}(\vect{V}) = \tfrac{1}{T-2}\!\sum_{t=1}^{T-2}\!\theta_t$
and its standard deviation $\phi_{\mathrm{ang}}(\vect{V}) =
\operatorname{std}\!\bigl(\{\theta_t\}_{t=1}^{T-2}\bigr)$, separating
consistent trajectories from erratic ones. Empirically, all four
\ours{}'s backbones straighten plausible
videos relative to violated ones (Appendix~\ref{app:straightening}).

\paragraph{Second-order motion: acceleration.}
\textit{A plausible sequence should also avoid abrupt changes in the feature-space displacement itself.} We capture this with the second-order difference $\vect{a}_t
    =
    \vect{v}_{t+1}
    -
    \vect{v}_t
    =
    \bar{\vect{z}}_{t+2}
    -
    2\bar{\vect{z}}_{t+1}
    +
    \bar{\vect{z}}_t \quad (t=1,\ldots,T-2)$.
This quantity is large when the trajectory abruptly changes speed or direction  (\textbf{(4)} ,$\vect{a}_t$ in Fig.~\ref{fig:curvature_illustration}). We summarize it as $\phi_{\mathrm{accel}}(\vect{V})
    =
    \frac{1}{T-2}
    \sum_{t=1}^{T-2}
    \|\vect{a}_t\|_2^2$. While $\phi_{\mathrm{curv}}$ isolates directional bending, $\phi_{\mathrm{accel}}$ captures the full second-order instability of the feature trajectory, including both changes in direction and changes in step magnitude.

\paragraph{Local linearity residual.}
A complementary signal, inspired by Goroshin et al.~\cite{goroshin2015linearize}'s
linearisation objective, asks whether the next feature lies in the affine
subspace spanned by the previous $H$. When local dynamics are linear, the
trajectory is locally low-dimensional: the next feature is a predictable
linear combination of its predecessors. Physical violations that break this
subspace (abruptly reversing fluids, spontaneously accelerating objects) spike
the residual at the violation frame. We fit a linear auto-regressive
predictor $\hat{P}_H\!:\!\mathbb{R}^{Hd}\!\to\!\mathbb{R}^{d}$ on past
windows, $\hat{\bar{\vect{z}}}_{t+1} = \hat{P}_H\!\bigl([\bar{\vect{z}}_{t-H+1};
\ldots;\bar{\vect{z}}_t]\bigr)$ for $t = H,\ldots,T{-}1$, and take the residual
$\vect{\varepsilon}_t = \bar{\vect{z}}_{t+1} - \hat{\bar{\vect{z}}}_{t+1}$
($[\cdot;\cdot]$: concatenation). Geometrically, $\vect{\varepsilon}_t$ is
the component of $\bar{\vect{z}}_{t+1}$ orthogonal to the $H$-step linear
span (\textbf{(3)}, $\varepsilon_t$ in Fig.~\ref{fig:curvature_illustration}).
We summarise it as $\phi_{\mathrm{perr}}(\vect{V}) = \tfrac{1}{T-H}
\sum_{t=H}^{T-1}\|\vect{\varepsilon}_t\|_2$; low values indicate a low-dimensional manifold.


\noindent\fbox{
  \begin{minipage}{\dimexpr\linewidth-2\fboxsep-2\fboxrule\relax}
    Together, these five scalar signals form the temporal geometric descriptor used by \ours{}:
    \begin{equation}
        \Phi_{\mathrm{temp}}(\vect{V})
        =
        \big(
        \phi_{\mathrm{curv}},
        \phi_{\mathrm{ang}},
        \phi_{\mathrm{speed}},
        \phi_{\mathrm{accel}},
        \phi_{\mathrm{perr}}
        \big).
        \label{eq:temporal_descriptor}
    \end{equation}
    For all five signals, larger values indicate less regular feature-space dynamics.
  \end{minipage}
}


\subsection{Vision backbones and \ours{} pipeline}
\label{sec:method_features}

%
%
\begin{figure}[!htb]
\centering
\resizebox{\textwidth}{!}{%
\begin{tikzpicture}[
    >=Stealth,
    every node/.style={inner sep=0pt, outer sep=0pt},
]

\node[coltitle] at (1.10, 2.55) {Input frames};
\node[coltitle] at (3.8,  2.55) {Frozen backbone $f_\theta$};
\node[coltitle] at (8.40, 2.55) {Architecture \& readout layer};
\node[coltitle] at (12.85,2.55) {Trajectory $\Gamma(\vect{V})$};

\draw[paleline, line width=0.4pt] (2.20, 0.00) -- (14.40, 0.00);
\node[font=\tiny\itshape\sffamily, text=black, anchor=west]
  at (14.80,  1) {\textsc{Self-supervised ViTs}};
\node[font=\tiny\itshape\sffamily, text=black, anchor=west]
  at (14.80, -1) {\textsc{Neuro-inspired models}};

\foreach \i/\y in {1/1.90, 2/0.95, 3/0.00, 4/-0.95, 5/-1.90}{
  \node[framenode, inner sep=0pt] (frame\i) at (1.10, \y)
    {\includegraphics[width=1.2cm, height=0.85cm]{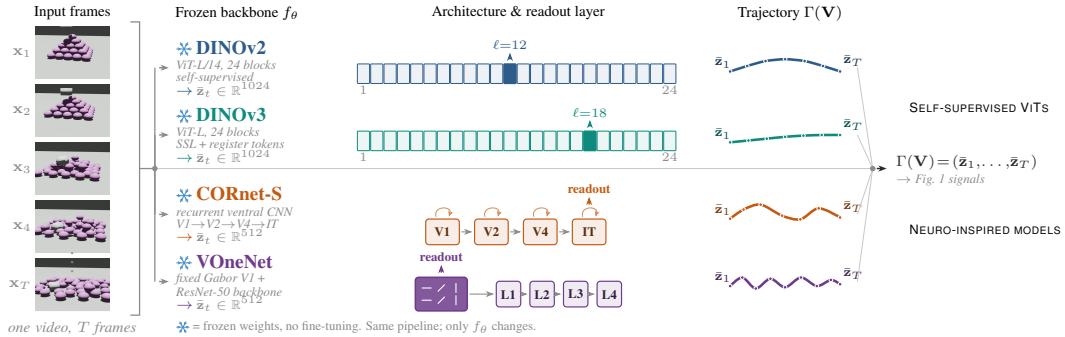}};
}
\foreach \i/\y/\lab in {1/1.90/{$\vect{x}_1$},   2/0.95/{$\vect{x}_2$},
                        3/0.00/{$\vect{x}_3$},   4/-0.95/{$\vect{x}_4$},
                        5/-1.90/{$\vect{x}_T$}}{
  \node[font=\scriptsize, text=softgrey, anchor=east, inner sep=2pt]
    at (frame\i.west) {\lab};
}
\node[font=\small, text=softgrey] at (1.10, -1.425) {$\vdots$};

\draw[softgrey, line width=0.6pt]
  (2.05, 2.40) -- (2.20, 2.40) -- (2.20, -2.40) -- (2.05, -2.40);

\node[font=\scriptsize\itshape, text=softgrey, anchor=north, text width=2.2cm, align=center]
  at (1.10, -2.50) {one video, $T$ frames};

\draw[softgrey, line width=0.6pt] (2.20, 0.00) -- (2.45, 0.00);
\fill[softgrey] (2.45, 0.00) circle (0.045);
\foreach \yy in {1.65, 0.55, -0.75, -1.85}{
  \draw[softgrey, line width=0.6pt] (2.45, 0.00) -- (2.45, \yy);
  \draw[fanoutarr] (2.45, \yy) -- (2.75, \yy);
}

\def\yA{1.65}
\node[anchor=west, font=\small] at (2.80, \yA+0.32)
  {\snowflake[0.12cm]\hspace{2pt}\textbf{\textcolor{ditchblue}{DINOv2}}};
\node[rowsub, anchor=west] at (2.80, \yA+0.04) {ViT-L/14, 24 blocks};
\node[rowsub, anchor=west] at (2.80, \yA-0.16) {self-supervised};
\node[rowdim, anchor=west] at (2.80, \yA-0.36)
  {{\scriptsize\color{ditchblue}$\to$} $\bar{\vect{z}}_t \in \mathbb{R}^{1024}$};
\node[anchor=center] at (8.40, \yA) {\vitstack{ditchblue}{12}};
\node[anchor=center] at (12.85, \yA+0.05) {\trajA{ditchblue}};

\def\yB{0.55}
\node[anchor=west, font=\small] at (2.80, \yB+0.32)
  {\snowflake[0.12cm]\hspace{2pt}\textbf{\textcolor{ditchteal}{DINOv3}}};
\node[rowsub, anchor=west] at (2.80, \yB+0.04) {ViT-L, 24 blocks};
\node[rowsub, anchor=west] at (2.80, \yB-0.16) {SSL\,+\,register tokens};
\node[rowdim, anchor=west] at (2.80, \yB-0.36)
  {{\scriptsize\color{ditchteal}$\to$} $\bar{\vect{z}}_t \in \mathbb{R}^{1024}$};
\node[anchor=center] at (8.40, \yB) {\vitstack{ditchteal}{18}};
\node[anchor=center] at (12.85, \yB+0.05) {\trajB{ditchteal}};

\def\yC{-0.75}
\node[anchor=west, font=\small] at (2.80, \yC+0.32)
  {\snowflake[0.12cm]\hspace{2pt}\textbf{\textcolor{cornorange}{CORnet-S}}};
\node[rowsub, anchor=west] at (2.80, \yC+0.04) {recurrent ventral CNN};
\node[rowsub, anchor=west] at (2.80, \yC-0.16) {V1$\to$V2$\to$V4$\to$IT};
\node[rowdim, anchor=west] at (2.80, \yC-0.36)
  {{\scriptsize\color{cornorange}$\to$} $\bar{\vect{z}}_t \in \mathbb{R}^{512}$};
\node[anchor=center] at (8.40, \yC) {\cornetschem{cornorange}{IT}};
\node[anchor=center] at (12.85, \yC+0.05) {\trajC{cornorange}};

\def\yD{-1.85}
\node[anchor=west, font=\small] at (2.80, \yD+0.32)
  {\snowflake[0.12cm]\hspace{2pt}\textbf{\textcolor{voneplum}{VOneNet}}};
\node[rowsub, anchor=west] at (2.80, \yD+0.04) {fixed Gabor V1\,+};
\node[rowsub, anchor=west] at (2.80, \yD-0.16) {ResNet-50 backbone};
\node[rowdim, anchor=west] at (2.80, \yD-0.36)
  {{\scriptsize\color{voneplum}$\to$} $\bar{\vect{z}}_t \in \mathbb{R}^{512}$};
\node[anchor=center] at (8.40, \yD) {\vonenetschem{voneplum}{vone}};
\node[anchor=center] at (12.85, \yD+0.05) {\trajD{voneplum}};

\foreach \yy in {1.65, 0.55, -0.75, -1.85}{
  \draw[paleline, line width=0.4pt] (13.95, \yy) -- (14.20, 0.0);
}
\fill[softgrey] (14.20, 0.0) circle (0.045);
\draw[fanoutarr, line width=0.7pt, draw=inkgrey]
  (14.20, 0.0) -- (14.50, 0.0);
\node[font=\scriptsize, text=inkgrey, anchor=west, inner sep=1pt]
  at (14.55, 0.10) {$\Gamma(\vect{V})\!=\!(\bar{\vect{z}}_1,\!\ldots,\!\bar{\vect{z}}_T)$};
\node[font=\tiny\itshape, text=softgrey, anchor=west, inner sep=1pt]
  at (14.55, -0.18) {$\to$ Fig.~1 signals};

\node[anchor=west, font=\tiny, text=softgrey] at (2.80, -2.60)
  {\snowflake[0.10cm]\ = frozen weights, no fine-tuning. Same pipeline; only $f_\theta$ changes.};

\end{tikzpicture}%
}

\caption{\textbf{\textsc{GeoPhys} pipeline.}
A single video $\vect{V}=(\vect{x}_1,\ldots,\vect{x}_T)$ feeds each
backbone $f_\theta$ unchanged.  Each backbone reads out at the layer
$\ell^\star$ selected on a held-out validation split
(Appendix~\ref{app:layer_sweep}); the per-frame embedding
$\bar{\vect{z}}_t = \mathrm{pool}\!\bigl(f_\theta^{(\ell^\star)}(\vect{x}_t)\bigr)$
is spatial-pooled and stacked across time into the trajectory
$\Gamma(\vect{V})=(\bar{\vect{z}}_1,\ldots,\bar{\vect{z}}_T)$, on
which the geometric signals of Fig.~\ref{fig:curvature_illustration}
operate.}
\label{fig:backbone_pipeline}
\end{figure}

\begin{figure}[t]
    \centering
    \includegraphics[width=\textwidth]{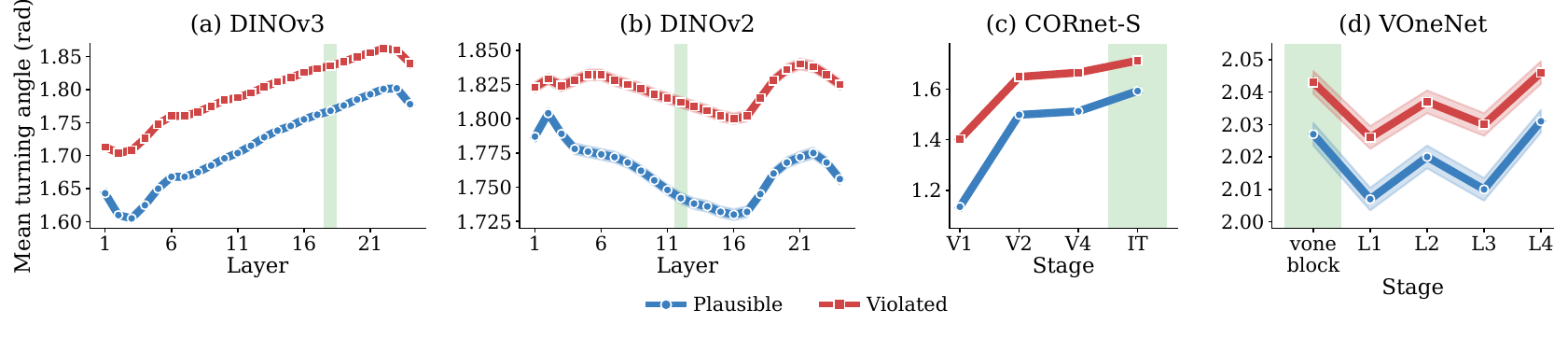}
    \caption{\textbf{Mean curvature across layers} Blue: plausible. Red:
    violated (65 video pairs). Violated lies
    above plausible at every layer of every backbone ($p < 10^{-8}$, $t$-test). Green band: readout layer.}
    \label{fig:app_straightening_layers}
\end{figure}

\paragraph{From frames to temporal feature trajectories.}
\ours{} operates on any frozen vision backbone without retraining or modification. Given a backbone $f_\theta$ and a held-out validation split, we select a readout layer $\ell^\star$ that maximizes the
curvature gap between plausible and violated videos (Fig.~\ref{fig:app_straightening_layers}), yielding the frozen
feature map $f(\cdot) = f_\theta^{(\ell^\star)}(\cdot)$. For a video
$\vect{V}=\{\vect{x}_1,\ldots,\vect{x}_T\}$, each frame produces $N$
spatial tokens of dimension $d$:
$Z_t = f(\vect{x}_t) = \{\vect{z}_{t,n}\}_{n=1}^{N} \in \mathbb{R}^{N \times d}$.
Spatial pooling collapses each grid into a single vector
$\bar{\vect{z}}_t = \mathrm{pool}(Z_t) \in \mathbb{R}^{d}$. Stacking
these across time forms the trajectory
$\Gamma(\vect{V}) = (\bar{\vect{z}}_1, \ldots, \bar{\vect{z}}_T)$,
upon which \ours{}'s signals
(Sec.~\ref{sec:method_signals}) operate. This pipeline remains
identical across all backbones, changing only the choice of $f_\theta$
(Fig.~\ref{fig:backbone_pipeline}).

\paragraph{Backbone selection.}
Hypothesis~\ref{box:hypothesis} originates in the V1-straightening
literature~\cite{henaff2019straightening,henaff2021straightening}, so
testing it requires backbones that span V1-like and non-V1-like
representations. We use four frozen backbones in two architecture
classes. \textbf{DINOv2}~\cite{oquab2024dinov2} and
\textbf{DINOv3}~\cite{simoni2025dinov3} are self-supervised ViT-L/14s
already shown to discriminate AI-generated from natural video via the
curvature signal~\cite{interno2026aigenerated}.
\textbf{CORnet-S}~\cite{kubilius2019cornet} (recurrent
V1$\to$V2$\to$V4$\to$IT ventral-stream CNN~\cite{kar2019recurrent}) and
\textbf{VOneNet}~\cite{vonenet2020} (ResNet-50 with a fixed Gabor V1
front-end~\cite{heeger1992lnp,carandini2012normalization}) are
architecturally explicit models of primate visual cortex.

\paragraph{Readout via curvature.}
For each backbone, we select the readout layer $\ell^\star$ that maximises
the curvature delta $\Delta\phi_{\mathrm{curv}} = \bar{\phi}_{\mathrm{curv}}^{-}
- \bar{\phi}_{\mathrm{curv}}^{+}$ between violated and plausible videos on
a held-out LikePhys validation split~\cite{yuan2025likephys}. By
Hypothesis~\ref{box:hypothesis}, $\ell^\star$ best reflects the distinction
\ours{} exploits. The geometric and task-driven criteria coincide: for
every backbone, $\ell^\star$ also maximises detection 
accuracy
(Appendix~\ref{app:layer_sweep}). The gap is positive across all $57$
layer$\times$backbone combinations ($p < 10^{-8}$, paired $t$-test;
$d \in [0.22, 0.58]$; Fig.~\ref{fig:app_straightening_layers}), with
CORnet-S~V1 producing the largest effect and DINOv2/3 rising with depth.
For backbone $b$ with readout $\ell^\star$, we write
$u^{(b)}_{\phi}(\vect{V}) = \phi\bigl(\Gamma_{f_\theta}(\vect{V})\bigr)$
for the value of signal $\phi$ on $\vect{V}$'s trajectory through that
backbone.



\paragraph{Decision rule.}
Each backbone uses the single signal that scores best on the LikePhys~\cite{yuan2025likephys} held-out
split: angle consistency for DINOv2/v3, speed variation for CORnet-S,
and acceleration for VOneNet (Table~\ref{tab:signals}). \ours{} maps a video
to a single scalar, its standardized signal $z_b$, with larger values
indicating less plausible dynamics. We predict the video with the larger $z_b$ as violated, and read
$|z_b|$ as confidence. Because the backbones catch different violations
(Fig.~\ref{fig:complementarity}), we combine them by \textbf{Majority}
($\geq 3/4$ agree) or \textbf{OR} (the most confident,
$b^\star=\arg\max_b|z_b|$).

\section{Evaluation}
\label{sec:method_applications}
We test \ours{} in three studies, applying the same score unchanged
throughout, to show that feature-space geometry reflects physical plausibility, that the resulting signal is biologically
meaningful, and that it is useful for physics-aligned video generation.
\textbf{Study~1 (Sec.~\ref{sec:neuro})} compares \ours{} signals to EEG visual working-memory recordings from humans viewing matched
violation-of-expectation stimuli~\cite{balaban2024electrophysiology,
liu2024violations, kaiser2023eeg}, testing whether the per-frame score
tracks neuro-biological responses.
\textbf{Study~2 (Sec.~\ref{sec:exp_detection})} discriminates plausible
from violated videos on LikePhys~\cite{yuan2025likephys} and
IntPhys2~\cite{bordes2025intphys2}.
\textbf{Study~3 (Sec.~\ref{sec:exp_bon})} deploys the score as a
training-free best-of-$N$ verifier on
PhysicsIQ~\cite{motamed2025physicsiq}.

\subsection{Study 1: Detecting object permanence violations in models and brains}\label{sec:neuro}
\begin{figure}[htb]
    \centering
    \includegraphics[width=1\linewidth]{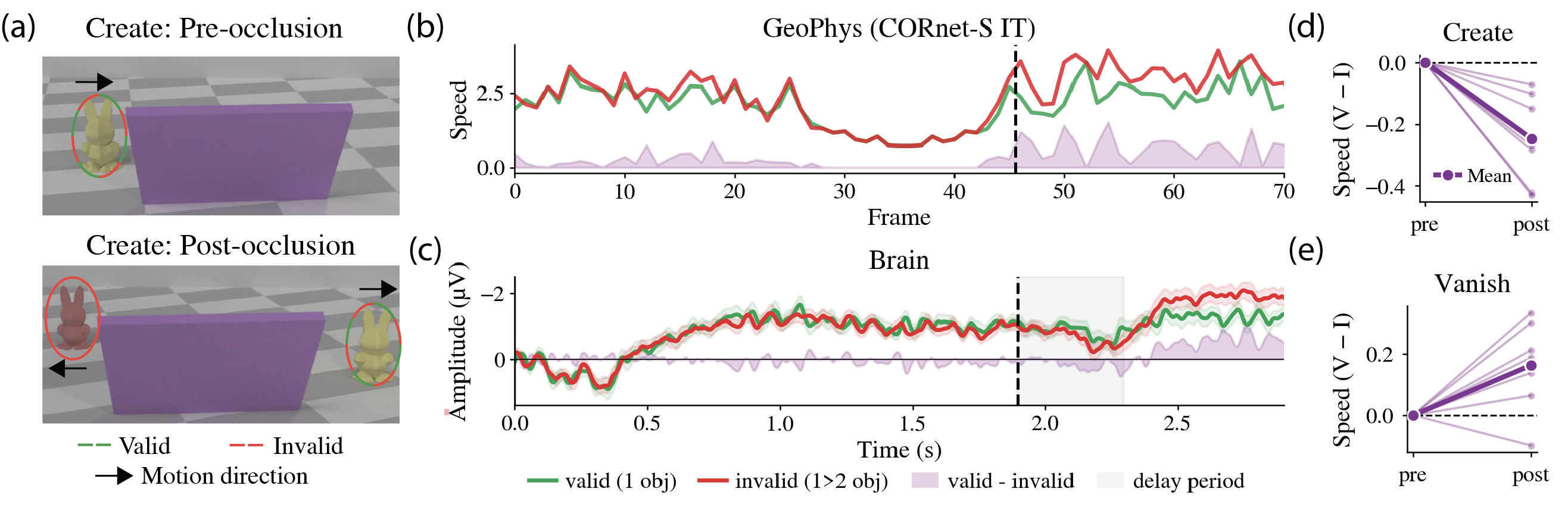}
    \caption{
        \textbf{Model and brain comparison.}
        (a) Example VOE stimuli for the Create scenario: in valid videos, one object enters and exits occlusion; in invalid videos, one enters but two exit. (b--c) Valid and invalid Create signals from \ours{} CORnet-S IT speed (b) and EEG contralateral delay activity (from \cite{balaban2024electrophysiology}; c). Both are elevated for the invalid condition after occlusion offset (vertical dashed line). (d--e) Mean valid $-$ invalid \ours{} CORnet-S IT speed pre- and post-occlusion for Create (d) and Vanish (e) scenarios (rendered with ADEPT). Known EEG VOE delay period in gray~\cite{balaban2024electrophysiology}.
    }
    \label{fig:4_Neuro}
\end{figure}

\paragraph{Setup.}
The violation-of-expectation (VOE) paradigm uses paired valid/invalid stimuli to elicit reliable surprise responses to physically impossible events in infants~\cite{margoni2024violation,dunn2017investigating} and, in adult EEG, rapid shifts in the contralateral delay activity (CDA), a marker of visual working memory that scales with the number of tracked objects~\cite{balaban2024electrophysiology,ikkai2010contralateral,mccollough2007electrophysiological}. Across two experiments, we test whether \ours{} per-frame signals $(\theta_t, s_t, \vect{a}_t, \vect{\varepsilon}_t)$ reliably track the same time course on two object-permanence scenarios: \emph{Create} (one object enters an occluder, two emerge) and \emph{Vanish} (two enter, one emerges). Experiment~1: we use the public stimulus subset from~\cite{balaban2024electrophysiology} ($n=16$ subjects) to visualize temporal alignment. Experiment~2: we render additional matched pairs with ADEPT~\cite{smith2019adept} and measure the valid $-$ invalid signal difference pre- and post-occlusion.

\paragraph{Findings.}
In Create (Fig.~\ref{fig:4_Neuro}a), \ours{} CORnet-S IT speed (Fig.~\ref{fig:4_Neuro}b) and the human CDA (Fig.~\ref{fig:4_Neuro}c) both rise rapidly and persistently in the invalid condition after the violation; in Vanish (Fig.~\ref{fig12: neuro_vanish_ts}a), both decrease. The two signals diverge during occlusion (${\sim}1.2$--$1.9$\,s), where the CDA stays elevated, consistent with abstract object tracking beyond the visual input~\cite{balaban2024electrophysiology}, whereas \ours{} approaches zero. Quantitatively, \ours{} tracks object number in Create ($t(6) = 4.501$, $p = 0.0041$; Fig.~\ref{fig:4_Neuro}d) and 
Vanish ($t(6) = -2.926$, $p = 0.0264$; Fig.~\ref{fig:4_Neuro}e).
Across backbones (Table~\ref{tab:prepost_transposed}), CORnet-S IT speed, acceleration, and prediction error are the only signals showing notable post-occlusion divergence across Create and Vanish. Per-backbone and ensemble detection on all stimuli is reported in Tab.~\ref{tab:heatmap_neuro}.

\begin{takeaway}
\ours{} signals align with human EEG responses to object-permanence violations~\cite{balaban2024electrophysiology} and scale with object number, consistent with physical-violation perception.
\end{takeaway}

\subsection{Study 2: Physics violation detection}
\label{sec:exp_detection}

\begin{figure}[htb]
\centering
\includegraphics[width=\textwidth]{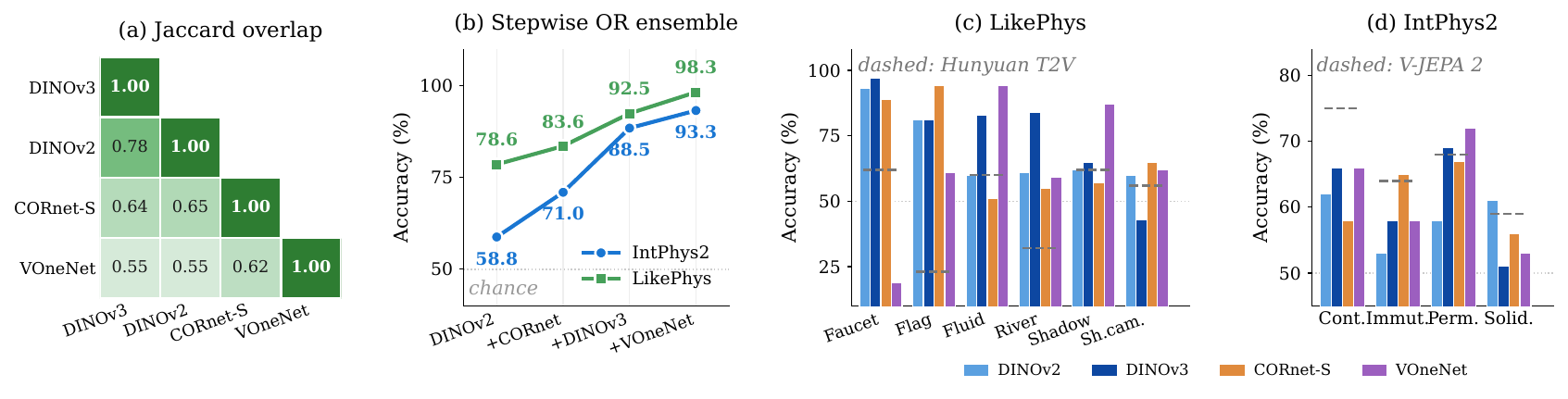}
\caption{\textbf{Backbones are complementary.}
\textbf{(a)}~Jaccard overlap of correctly-flagged sets on LikePhys.
\textbf{(b)}~Greedy stepwise ensemble reaches $98.3\%$ on LikePhys
and $93.3\%$ on IntPhys2.
\textbf{(c)}~Each backbone wins a different
physics domain of LikePhys.
\textbf{(d)}~Each backbone wins a different
condition in IntPhys2. Dashed: best baseline (Hunyuan T2V~\cite{hunyuan2024}
in c, V-JEPA~2~\cite{assran2025vjepa2} in d).}
\label{fig:complementarity}
\end{figure}

 \newcommand{\scenthumb}[3]{%
  \begin{minipage}[b]{#1}%
    \centering
    \includegraphics[width=\linewidth]{figures/thumbs/#2}\\[-1pt]
    {\fontsize{6pt}{6.8pt}\selectfont #3}%
  \end{minipage}%
}
\newcommand{\likephysrowone}{%
  \scenthumb{9mm}{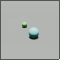}{Ball coll.}\hfill%
  \scenthumb{9mm}{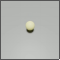}{Ball drop}\hfill%
  \scenthumb{9mm}{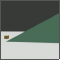}{Block sl.}\hfill%
  \scenthumb{9mm}{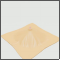}{Cloth}\hfill%
  \scenthumb{9mm}{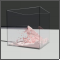}{Faucet}\hfill%
  \scenthumb{9mm}{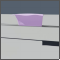}{Flag}%
}
\newcommand{\likephysrowtwo}{%
  \scenthumb{9mm}{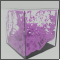}{Fluid}\hfill%
  \scenthumb{9mm}{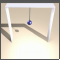}{Pendulum}\hfill%
  \scenthumb{9mm}{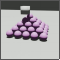}{Pyramid}\hfill%
  \scenthumb{9mm}{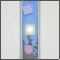}{River}\hfill%
  \scenthumb{9mm}{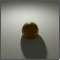}{Shadow}\hfill%
  \scenthumb{9mm}{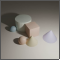}{Sh.\ cam.}%
}
\renewcommand{\scenthumb}[3]{%
  \begin{minipage}[b]{#1}%
    \centering
    \includegraphics[width=\linewidth]{figures/thumbs/#2}\\[1pt]
    {\fontsize{15pt}{9pt}\selectfont #3}%
  \end{minipage}%
}

\definecolor{hm100}{RGB}{144,197,148} 
\definecolor{hm90}{RGB}{180,215,182}
\definecolor{hm80}{RGB}{205,230,207}
\definecolor{hm70}{RGB}{230,242,231}
\definecolor{hm60}{RGB}{248,250,248}  
\definecolor{hm45}{RGB}{250,220,220}  
\definecolor{hm19}{RGB}{240,170,175}  

\newcommand{\hmg}[1]{\cellcolor{hm100}\textcolor{black}{\textbf{#1}}}
\newcommand{\hmG}[1]{\cellcolor{hm90}\textcolor{black}{#1}}
\newcommand{\hmGl}[1]{\cellcolor{hm80}\textcolor{black}{#1}}
\newcommand{\hmGll}[1]{\cellcolor{hm70}\textcolor{black}{#1}}
\newcommand{\hmW}[1]{\cellcolor{hm60}\textcolor{black}{#1}}
\newcommand{\hmRl}[1]{\cellcolor{hm45}\textcolor{black}{#1}}
\newcommand{\hmR}[1]{\cellcolor{hm19}\textcolor{black}{\textbf{#1}}}
\newcommand{\hmgb}[1]{\cellcolor{hm100}\textcolor{black}{\textbf{#1}}}
\newcommand{\hmGb}[1]{\cellcolor{hm90}\textcolor{black}{\textbf{#1}}}
\newcommand{\hmGlb}[1]{\cellcolor{hm80}\textcolor{black}{\textbf{#1}}}
\newcommand{\hmGllb}[1]{\cellcolor{hm70}\textcolor{black}{\textbf{#1}}}
\newcommand{\hmWb}[1]{\cellcolor{hm60}\textcolor{black}{\textbf{#1}}}
\paragraph{Setup.}
LikePhys~\cite{yuan2025likephys} and IntPhys2~\cite{bordes2025intphys2}
present matched pairs $(\vect{V}^+, \vect{V}^-)$ sharing initial
conditions, with one physically violated; the task is to identify it.
The per-backbone scalar is the signed delta
$\delta_b = u^{(b)}_{\phi_b}(\vect{V}^-) - u^{(b)}_{\phi_b}(\vect{V}^+)$;
we z-normalise it to $z_b$, whose sign identifies the violated video, and
aggregate by Majority or OR exactly as defined in
Sec.~\ref{sec:method_features}. $95\%$ CIs: $1$K-resample bootstrap
(LikePhys~\cite{yuan2025likephys} per-scenario, IntPhys2~\cite{bordes2025intphys2} per-pair).

\begin{table*}[htb]
\caption{\textbf{Physics violation detection.}
Pairwise accuracy (\%, $\uparrow$ 0\%
\protect\tikz[baseline=-0.5ex]{
  \protect\shade[left color=hm19, right color=hm60] (0,0) rectangle (0.75,0.2);
  \protect\shade[left color=hm60, right color=hm100] (0.75,0) rectangle (1.5,0.2);
} 100\%).
\underline{Underline}: best baseline per column.
\textbf{Bold}: best \ours{} backbone per column.}
\label{tab:detection}
\centering
\renewcommand{\scenthumb}[3]{%
  \begin{tabular}[b]{@{}c@{}}
    \fontsize{20pt}{12pt}\selectfont\rotatebox{40}{#3}\\[2pt]
    \includegraphics[width=#1]{figures/thumbs/#2}
  \end{tabular}%
}

\resizebox{\textwidth}{!}{%
\renewcommand{\arraystretch}{1.25}%
\setlength{\tabcolsep}{3pt}%
\LARGE
\begin{tabular}{l cccccccccccc c | l cccc c}
\toprule
& \multicolumn{13}{c|}{\huge\textbf{LikePhys}~\cite{yuan2025likephys}}
& \multicolumn{6}{c}{\huge\textbf{IntPhys2}~\cite{bordes2025intphys2}}\\
& \scenthumb{15mm}{lp_ball_collision}{Ball coll.}
& \scenthumb{15mm}{lp_ball_drop}{Ball drop}
& \scenthumb{15mm}{lp_block_slide}{Block slide}
& \scenthumb{15mm}{lp_cloth_drape}{Cloth drape}
& \scenthumb{15mm}{lp_faucet}{Faucet}
& \scenthumb{15mm}{lp_flag}{Flag}
& \scenthumb{15mm}{lp_fluid}{Fluid}
& \scenthumb{15mm}{lp_pendulum}{Pendulum}
& \scenthumb{15mm}{lp_pyramid}{Pyramid}
& \scenthumb{15mm}{lp_river}{River}
& \scenthumb{15mm}{lp_shadow}{Shadow}
& \scenthumb{15mm}{lp_shadow_camera}{Sh.\ cam.}
& \rotatebox{70}{M.Avg.}
&
& \scenthumb{15mm}{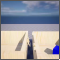}{Contin.}
& \scenthumb{15mm}{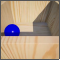}{Immut.}
& \scenthumb{15mm}{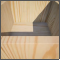}{Perman.}
& \scenthumb{15mm}{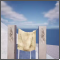}{Solid.}
& \rotatebox{70}{M.Avg.} \\
\midrule
AnimDiff~\cite{guo2024animatediff}
  & \hmR{37} & \hmRl{40} & \hmR{35} & \hmRl{47} & \hmR{38} & \hmR{21}
  & \hmR{37} & \hmRl{48} & \hmR{39} & \hmW{56}  & \hmR{28} & \hmRl{44}
  & 39.2
  & Cosmos~\cite{nvidia2025cosmosworldfoundationmodel}             & \hmRl{54} & \hmRl{51} & \hmRl{52} & \hmW{55} & 49.4 \\
AnimSDXL~\cite{guo2024animatediff}
  & \hmR{37} & \hmR{33} & \hmW{62} & \hmRl{53} & \hmRl{46} & \hmR{17}
  & \hmRl{40} & \hmR{30} & \hmR{31} & \hmRl{54} & \hmW{57} & \hmGll{68}
  & 44.0
  & Qwen-VL~\cite{yang2025qwen3technicalreport}            & \hmRl{54} & \hmW{57}  & \hmRl{54} & \hmRl{51} & 52.3 \\
ZeroScope~\cite{wang2023modelscopetexttovideotechnicalreport}
  & \hmRl{47} & \hmRl{45} & \hmRl{53} & \hmW{61} & \hmRl{54} & \hmR{16}
  & \hmR{38}  & \hmRl{42} & \hmR{27}  & \hmW{60} & \hmW{60}  & \hmW{58}
  & 46.7
  & Gemini 1.5~\cite{geminiteam2025geminifamilyhighlycapable}       & \hmW{55}  & \hmW{57}  & \hmW{56}  & \hmW{57}  & 52.3 \\
ModelScope~\cite{wang2023modelscopetexttovideotechnicalreport}
  & \hmRl{48} & \hmRl{47} & \hmRl{53} & \hmW{60} & \hmW{60}  & \hmR{16}
  & \hmR{38}  & \hmR{33}  & \hmR{21}  & \hmW{60} & \hmGll{67}& \hmW{62}
  & 47.1
  & GPT-4o~\cite{openai2024gpt4technicalreport}              & \hmW{57}  & \hmW{60}  & \hmW{60}  & \hmW{57}  & 53.8 \\
Mochi~\cite{mochi2024}
  & \hmW{60}  & \hmRl{50} & \hmGll{68}& \hmGll{71}& \hmW{62} & \hmR{17}
  & \hmRl{40} & \hmR{35}  & \hmR{16}  & \hmW{56} & \hmRl{42} & \hmW{60}
  & 48.1
  & VidMAEv2~\cite{videomaev2}       & \hmGll{65}& \hmW{55}  & \hmW{64}  & \hmW{60}  & 53.8 \\
CogVidX-5B~\cite{cogvideox2024}
  & \hmW{57}  & \hmW{58}  & \hmR{38}  & \hmW{61} & \hmW{64}  & \hmR{19}
  & \hmRl{47} & \hmR{35}  & \hmR{17}  & \hmW{60} & \hmGl{77} & \hmGll{70}
  & 50.2
  & V-JEPA~\cite{assran2025vjepa2}              & \hmW{57}  & \hmW{56}  & \hmGll{\underline{68}} & \hmRl{52} & 53.8 \\
CogVidX-2B~\cite{cogvideox2024}
  & \hmW{60}  & \hmW{62}  & \hmRl{40} & \hmW{64} & \hmW{64}  & \hmR{19}
  & \hmRl{50} & \hmR{37}  & \hmR{17}  & \hmW{60} & \hmGl{\underline{82}} & \hmGll{\underline{68}}
  & 51.8
  & Gemini 2.5~\cite{geminiteam2025geminifamilyhighlycapable}       & \hmW{55}  & \hmW{\underline{64}} & \hmW{64}  & \hmW{56}  & 55.6 \\
Wan2.1-1B~\cite{wan2025}
  & \hmW{57}  & \hmRl{47} & \hmR{33}  & \hmR{34} & \hmRl{40} & \hmGll{67}
  & \hmGll{67}& \hmGl{\underline{80}}& \hmW{60} & \hmR{22} & \hmGl{77} & \hmRl{40}
  & 52.0
  & V-JEPA 2~\cite{assran2025vjepa2} & \hmGll{\underline{75}} & \hmW{59} & \hmW{60}  & \hmW{\underline{59}} & 57.5 \\
LTX Vid.~\cite{hacohen2024ltxvideorealtimevideolatent}
  & \hmW{63}  & \hmRl{42} & \hmGll{65}& \hmRl{53}& \hmGll{\underline{68}} & \hmRl{51}
  & \hmGll{72}& \hmR{32}  & \hmGl{\underline{81}} & \hmW{60} & \hmR{25} & \hmRl{52}
  & 55.3
  & & & & & \\
CogVidX1.5~\cite{cogvideox2024}
  & \hmR{38}  & \hmRl{50} & \hmRl{48} & \hmRl{47}& \hmRl{46} & \hmGl{\underline{77}}
  & \hmGl{\underline{78}} & \hmGll{68}& \hmGll{74}& \hmR{32} & \hmGl{77}& \hmR{38}
  & 56.2
  & & & & & \\
Wan2.1-14B~\cite{wan2025}
  & \hmW{58}  & \hmRl{43} & \hmR{38}  & \hmR{37} & \hmRl{44} & \hmGll{66}
  & \hmG{85}  & \hmG{\underline{87}}& \hmW{57} & \hmR{36} & \hmGl{\underline{83}} & \hmRl{40}
  & 56.2
  & & & & & \\
Hunyuan~\cite{hunyuan2024}
  & \hmGll{\underline{67}} & \hmRl{\underline{52}} & \hmGl{\underline{75}} & \hmW{\underline{64}} & \hmW{62} & \hmR{23}
  & \hmW{60}  & \hmGl{78} & \hmRl{50} & \hmR{32} & \hmW{62}  & \hmW{56}
  & 56.4
  & & & & & \\
\midrule
\multicolumn{20}{l}{\textit{$\dagger$: trained with leave-one-scene-out cross-validation on same features of \ours{} (Dinov3)}} \\
Lin.\ probe$^\dagger$
  & \hmG{85} & \hmR{38} & \hmG{85} & \hmGl{75} & \hmGl{75} & \hmGll{65}
  & \hmRl{45} & \hmW{61} & \hmG{85} & \hmRl{45} & \hmRl{45} & \hmRl{45}
  & 62.4\,{$\pm$7.1}
  & Lin.\ probe$^\dagger$ & \hmRl{51} & \hmW{60} & \hmW{59} & \hmRl{53} & 55.5\,{$\pm$4.1} \\
\ours{} (V-JEPA 2)
  & \hmG{92} & \hmGll{68} & \hmGll{72} & \hmGl{75} & \hmG{92} & \hmGl{76}
  & \hmG{93} & \hmGl{81} & \hmg{97} & \hmW{66} & \hmGll{72} & \hmRl{57}
  & 78.3\,{$\pm$8.4}
  & \ours{} (V-JEPA 2) & \hmGll{67} & \hmW{58} & \hmW{60} & \hmRl{53} & 59.5\,{$\pm$5.5} \\
\midrule
\multicolumn{20}{l}{\textbf{\ours{} individual}} \\
DINOv2~\cite{oquab2024dinov2}
  & \hmgb{100}& \hmRl{45} & \hmgb{100}& \hmG{92} & \hmG{93}  & \hmGl{81}
  & \hmW{60}  & \hmG{89}  & \hmgb{100}& \hmW{61} & \hmW{62}  & \hmW{60}
  & 78.6\,{$\pm$8.2}
  & DINOv2~\cite{oquab2024dinov2}  & \hmW{62}  & \hmRl{53} & \hmW{58} & \hmWb{61}
  & 58.8\,{$\pm$7.1} \\
DINOv3~\cite{simoni2025dinov3}
  & \hmg{100} & \hmR{38}  & \hmg{100} & \hmG{92} & \hmGb{97} & \hmGl{81}
  & \hmGl{83} & \hmG{87}  & \hmg{100} & \hmGlb{84}& \hmGll{65}& \hmRl{43}
  & 80.8\,{$\pm$9.7}
  & DINOv3~\cite{simoni2025dinov3}  & \hmGllb{66}& \hmW{58} & \hmGll{69}& \hmRl{51}
  & 60.5\,{$\pm$4.5} \\
CORnet-S~\cite{kubilius2019cornet}
  & \hmGl{80} & \hmGlb{83}& \hmg{100} & \hmGl{78}& \hmG{89}  & \hmGb{94}
  & \hmRl{51} & \hmG{86}  & \hmg{100} & \hmW{55} & \hmW{57}  & \hmGllb{65}
  & 78.2\,{$\pm$9.2}
  & CORnet-S~\cite{kubilius2019cornet}& \hmW{58}  & \hmGllb{65}& \hmGll{67}& \hmW{56}
  & 61.1\,{$\pm$4.6} \\
VOneNet~\cite{vonenet2020}
  & \hmG{90}  & \hmGl{82} & \hmg{100} & \hmGb{88}& \hmR{19}  & \hmW{61}
  & \hmGb{94} & \hmGb{90} & \hmg{100} & \hmW{59} & \hmGb{87} & \hmW{62}
  & 77.6\,{$\pm$9.0}
  & VOneNet~\cite{vonenet2020} & \hmGll{66}& \hmW{58} & \hmGllb{72}& \hmRl{53}
  & \textbf{61.7}\,{$\pm$4.2} \\
\midrule
\multicolumn{20}{l}{\textbf{\ours{} ensembles}} \\
Majority
  & \hmg{100} & \hmGl{77} & \hmg{100} & \hmg{95} & \hmg{99}  & \hmg{96}
  & \hmG{93}  & \hmg{96}  & \hmg{100} & \hmG{86} & \hmGl{83} & \hmGl{78}
  & 90.9\,{$\pm$6.3}
  & Majority    & \hmGl{78} & \hmGl{75} & \hmGl{83}& \hmGl{75}& 77.5\,{$\pm$3.7} \\
OR
  & \hmg{100} & \hmG{88}  & \hmg{100} & \hmg{100}& \hmg{100} & \hmg{100}
  & \hmg{97}  & \hmg{99}  & \hmg{100} & \hmg{100}& \hmg{97}  & \hmG{92}
  & \textbf{98.3}\,{$\pm$2.1}
  & OR          & \hmG{94}  & \hmG{91}  & \hmG{95} & \hmG{93} & \textbf{93.3}\,{$\pm$2.8} \\
\midrule
  & & & & & & & & & & & & &
  & Human       & --        & --        & --       & --       & 96.4 \\
\bottomrule
\end{tabular}%
}
{\footnotesize\textit{IntPhys2 baselines: per-subset best runs (Bordes et al.~\cite{bordes2025intphys2}, Tab.~3); M.Avg.\ = their Main-set overall (Tab.~2).}}
\end{table*}

\paragraph{Data:}\textbf{LikePhys}~\cite{yuan2025likephys} provides 650 matched pairs rendered in Blender~\cite{Blender2018} across 12 scenarios in four domains (rigid-body, continuum, fluid, optical). Each pair
shares appearance and differs only by a controlled violation (reversed gravity, ground penetration, energy non-conservation, temporal disorder), probing whether models encode physical dynamics.
\textbf{IntPhys2}~\cite{bordes2025intphys2}, tests four core-knowledge properties (\emph{permanence},
\emph{solidity}, \emph{continuity},
\emph{immutability}) via 506 photorealistic 3D pairs in Unreal
Engine~\cite{EpicContentEULA}, with fixed and moving cameras and
structured as quadruplets where possible/impossible videos share
initial conditions and swap roles across occluder configurations.

\paragraph{Findings:}
\begin{wrapfigure}{r}{0.25\textwidth}
    \vspace{-1em}
    \centering
    \includegraphics[width=\linewidth]{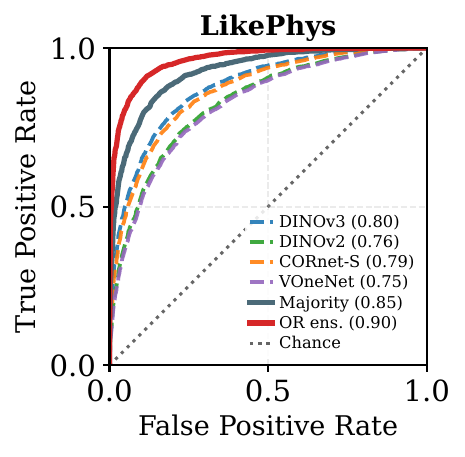}\\
    \includegraphics[width=\linewidth]{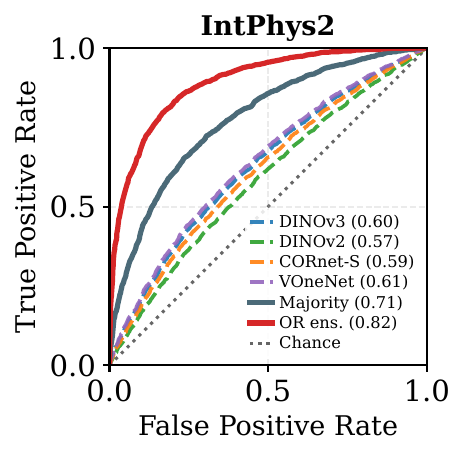}
    \caption{\textbf{ROC for physics violation detection.}
    LikePhys (top) and IntPhys2 (bottom).}
    \label{fig:roc_detection}
    \vspace{-3em}
\end{wrapfigure}
\textbf{LikePhys}~\cite{yuan2025likephys} ranks matched pairs via video
diffusion models' (DMs) likelihood preference: a DM judges plausibility
by assigning higher likelihood to the plausible video. All 12 DMs
evaluated score $39$--$56\%$ (near chance, Table~\ref{tab:detection},
left). A leave-one-scene-out linear probe trained on DINOv3 features
reaches $62.4\%$ (Table, $\dagger$). This probe controls for feature
quality. 
\ours{} gain is attributable to
trajectory geometry, not the backbone alone. Every \ours{} backbone
exceeds both: DINOv3~L18 $80.8\%$ (angle consistency), DINOv2~L12
$78.6\%$, CORnet-S~IT $78.2\%$ (speed variation), VOneNet~V1 $77.6\%$
(acceleration). 
Applying the same framework to V-JEPA~2 features yields
$78.3\%$, showing \ours{} is not backbone-specific. Each backbone
specialises in a different signal (full breakdown in
Appendix~\ref{app:signals}); majority vote reaches $90.9\%\!\pm\!6.3$
and OR $98.3\%\!\pm\!2.1$, within $1.7$ points of the per-scenario
ceiling (per-domain breakdown in App.~\ref{app:detection_analysis}). 

In \textbf{IntPhys2}~\cite{bordes2025intphys2}, V-JEPA~2
($57.5\%$), Gemini-2.5~Flash ($55.6\%$), GPT-4o ($53.8\%$), and a linear probe ($55.5\%$, Table, $\dagger$) all
remain within $8$ points of chance (Table~\ref{tab:detection}, right).
Every \ours{} backbone exceeds prior SoTA: VOneNet~V1 61.7\%, CORnet-S~V1 61.1\%, DINOv3~L12
60.5\%, DINOv2~L12 58.8\%. Majority vote reaches 77.5\%
($+20$ over V-JEPA~2) and OR reaches $93.3\%$, within $3.1$ points of
human performance ($96.4\%$); per-difficulty
and per-camera splits are in App.~\ref{app:detection_analysis}.

\vspace{-0.5em}
\paragraph{ROC analysis.}
Pairwise accuracy tests within-pair ranking only. ROC analysis
(Fig.~\ref{fig:roc_detection}) additionally tests cross-pair ranking
under a global threshold on the continuous score: $z_b$ for each
individual backbone, $\sum_b z_b$ for Majority, and $\max_b |z_b|$
for OR. Smooth curves at every operating point indicate the score is aligned with implausibility, not locally fitted to the
pairwise rule. AUCs reach $75$--$80\%$ on LikePhys (Majority $85\%$,
OR $90\%$) and $57$--$61\%$ on IntPhys2 (Majority $71\%$, OR $82\%$);
the ensemble lift over the best single backbone is $+10$ points on
LikePhys and $+21$ points on IntPhys2.

\vspace{-0.5em}
\paragraph{Backbone analysis.}
\label{sec:complementarity}
Each backbone covers a different set of violations
(Fig.~\ref{fig:complementarity}). Stepwise addition lifts the OR
ensemble to $98.3\%$ on LikePhys (from $78.6\%$) and $93.3\%$ on
IntPhys2 (from $58.8\%$). On LikePhys, CORnet-S wins flag, VOneNet wins fluid and shadow, and DINOv2/v3 capture faucet and river, exceeding $70$ points on a single scenario (Faucet: DINOv3 $97$ vs VOneNet $19$).
On IntPhys2, V1-like layers preserve signal
while DINOs require a mid-layer; mid-level features beat late ones
across all backbones (Appendix~\ref{app:layer_sweep}). The per-condition winners follow: VOneNet on permanence ($+4$ over
V-JEPA), CORnet-S on immutability ($+1$ over Gemini~2.5), DINOv2 on
solidity ($+2$ over V-JEPA~2).
Only on continuity does V-JEPA~2 give an edge, but majority vote
($78\%$) and OR ($94\%$) recover better performance.

\begin{takeaway}
\ours{} unlocks physics plausibility detection: every backbone surpasses all SoTA baselines on both benchmarks (LikePhys~\cite{yuan2025likephys}~$98\%$ \& IntPhys2~\cite{bordes2025intphys2}~$93\%$).
\end{takeaway}

\subsection{Study 3: Improved Physically Plausible Video Generation}
\label{sec:exp_bon}

\begin{figure}[h]
\centering
{\footnotesize\textbf{(A)}~PhysicsIQ scenarios}\par\vspace{1pt}
\setlength{\tabcolsep}{1pt}
\renewcommand{\arraystretch}{0.5}
\begin{tabular}{@{}ccccc@{}}
  \includegraphics[width=0.19\linewidth]{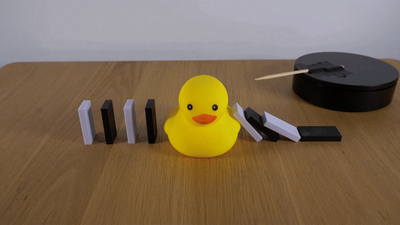} &
  \includegraphics[width=0.19\linewidth]{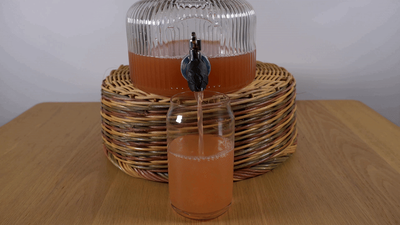} &
  \includegraphics[width=0.19\linewidth]{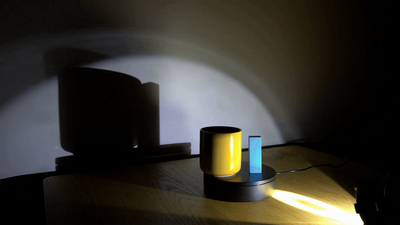} &
  \includegraphics[width=0.19\linewidth]{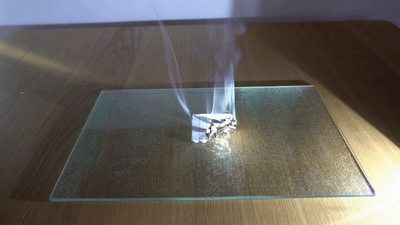} &
  \includegraphics[width=0.19\linewidth]{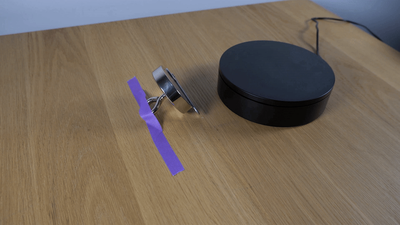} \\
  {\fontsize{6}{7}\selectfont Solid mechanics} &
  {\fontsize{6}{7}\selectfont Fluid dynamics} &
  {\fontsize{6}{7}\selectfont Optics} &
  {\fontsize{6}{7}\selectfont Thermo.} &
  {\fontsize{6}{7}\selectfont Magnetism} \\
\end{tabular}

\vspace{10pt}
{\footnotesize\textbf{(B)}~PhysicsIQ score per verifier, across four generators}
\includegraphics[width=\linewidth]{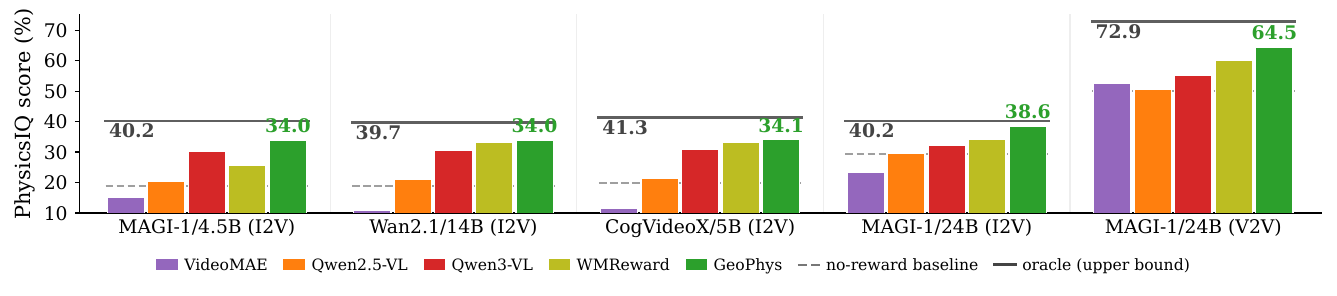}

\vspace{-4pt}
\caption{\textbf{PhysicsIQ benchmark and \ours{} performance.}
\textbf{(A)}~The five PhysicsIQ scenario categories.
\textbf{(B)}~Best-of-$N$ ($N{=}16$) PhysicsIQ score for five verifiers
on five generators; dashed strokes: no-verifier baseline, solid: oracle
upper bound.}
\label{fig:study3_overview}
\vspace{-1em}
\end{figure}

\begin{wrapfigure}{r}{0.45\textwidth}
    \vspace{-1.5em} 
    \centering
    \includegraphics[width=\linewidth]{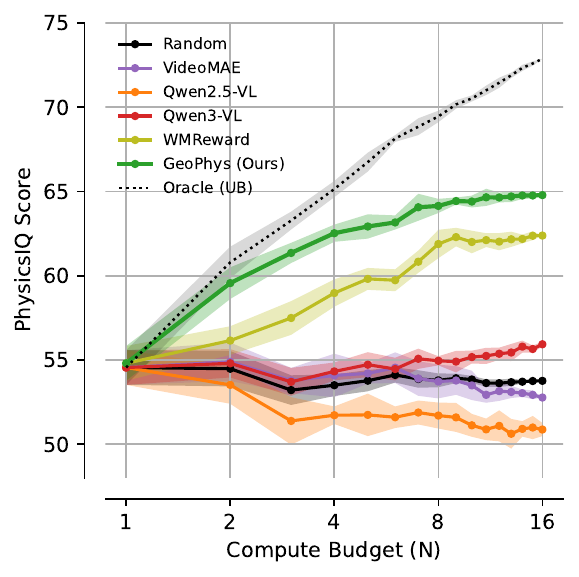}
    \caption{\textbf{Test-time scaling on MAGI-1 24B (V2V).} PhysicsIQ score as a function of the candidate budget $N \in \{1,\ldots, 16\}$. \ours{} scales closest to oracle.}
    \label{fig:v2v_scaling}
    \vspace{-2em} 
\end{wrapfigure}

\paragraph{Setup.}
We follow the test-time scaling setup of~\cite{liu2025video}: a
generator samples candidates, a verifier ranks them, and an oracle
bounds the achievable ceiling. We test all three on
PhysicsIQ~\cite{motamed2025physicsiq}, which provides $198$ scenarios,
each with a $3$\,s conditioning video and a held-out $5$\,s ground
truth. Outputs are scored by four motion-mask metrics aggregated into
a single PhysicsIQ score (Appendix~\ref{app:physicsiq}). For each
generator, we sample $N{=}16$ candidates per scenario. The verifier
selects one. \ours{} reuses the per-backbone signals and OR rule defined in
Sec.~\ref{sec:method_features} unchanged.
We evaluate four I2V generators
(MAGI-1~4.5B distill~\cite{magi1},
CogVideoX-5B~\cite{cogvideox2024}, Wan2.1~14B~\cite{wan2025},
MAGI-1~24B) and one V2V setting (MAGI-1~24B). For scaling curves, we
re-rank random subsets of size $N$ from the same pool
(Fig.~\ref{fig:v2v_scaling}). $N{=}1$ is no-reranking, reported as mean$\pm$std across $198$ scenarios.

\paragraph{Baselines.}
We compare four inference-time BoN baselines.
\textbf{VideoMAE(BoN)}~\cite{videomaev2} uses pixel-space
reconstruction error on masked spatiotemporal patches as a surprise
score. \textbf{Qwen2.5-VL(BoN)} and
\textbf{Qwen3-VL(BoN)}~\cite{yang2025qwen3technicalreport} prompt a multimodal-LLM with a
binary physics-plausibility question and use the positive-token logit
as the score.
\textbf{WMReward(BoN)}~\cite{yuan2026wmreward} uses V-JEPA~2 latent
prediction error~\cite{assran2025vjepa2}, a $1.1$B-parameter
video-pretrained world model. We also report  (supervised) \emph{Oracle} upper
bound: the candidate with the highest PhysicsIQ score per video.

\begin{table*}[t]
\caption{\textbf{PhysicsIQ results} ($N{=}16$).
For each generator, four motion-mask metrics (Sp.~IoU, Sp-Temp.~IoU,
W.~Sp.~IoU, MSE) and PhysicsIQ score (\%);
$\Delta$ over the baseline
(\protect\tikz[baseline=-0.5ex]{
  \protect\shade[left color=hm19, right color=hm60] (0,0) rectangle (0.6,0.18);
  \protect\shade[left color=hm60, right color=hm100] (0.6,0) rectangle (1.2,0.18);
}).}
\label{tab:bon_full}
\centering
\setlength{\tabcolsep}{3pt}
\setlength{\aboverulesep}{1pt}
\setlength{\belowrulesep}{1pt}

\vspace{-4pt}
\resizebox{\textwidth}{!}{%
\small
\begin{tabular}{l ccccc | ccccc | ccccc}
\toprule
& \multicolumn{5}{c|}{\textbf{MAGI-1 4.5B (I2V)}}
& \multicolumn{5}{c|}{\textbf{Wan2.1 14B (I2V)}}
& \multicolumn{5}{c}{\textbf{CogVideoX-5B (I2V)}}\\
\cmidrule(lr){2-6}\cmidrule(lr){7-11}\cmidrule(lr){12-16}
Verifier
& Sp & SpT & W.Sp & MSE & PhyIQ
& Sp & SpT & W.Sp & MSE & PhyIQ
& Sp & SpT & W.Sp & MSE & PhyIQ \\
\midrule
\textbf{Baseline}
  & .143 & .133 & .070 & .019 & 18.8
  & .143 & .133 & .070 & .018 & 18.9
  & .143 & .133 & .070 & .008 & 19.9 \\
+ VideoMAE
  & .122 & .044 & .042 & .029 & \hmRl{15.0}
  & .122 & .044 & .042 & .013 & \hmRl{11.0}
  & .122 & .044 & .042 & .008 & \hmRl{11.4} \\
+ Qwen2.5-VL
  & .168 & .132 & .072 & .018 & \hmW{20.3}
  & .168 & .132 & .072 & .011 & \hmW{21.0}
  & .168 & .132 & .072 & .008 & \hmW{21.3} \\
+ Qwen3-VL
  & .227 & .193 & .117 & .015 & \hmG{30.1}
  & .227 & .193 & .117 & .009 & \hmG{30.7}
  & .227 & .193 & .117 & .007 & \hmG{30.9} \\
+ WMReward~\cite{yuan2026wmreward}
  & .148 & .205 & .082 & .012 & \hmGl{25.6}
  & .235 & .210 & .130 & .008 & \hmG{33.1}
  & .235 & .210 & .130 & .007 & \hmG{33.2} \\
\best{+ \ours{} (Ours)}
  & \best{.225} & \best{.232} & \best{.128} & \best{.007} & \hmg{\textbf{34.0}}
  & \best{.225} & \best{.232} & \best{.128} & \best{.007} & \hmg{\textbf{34.0}}
  & \best{.225} & \best{.232} & \best{.128} & \best{.006} & \hmg{\textbf{34.1}} \\
\textit{Oracle (Upper bound)}
  & \textit{.304} & \textit{.245} & \textit{.161} & \textit{.017} & \textit{40.2}
  & \textit{.297} & \textit{.236} & \textit{.158} & \textit{.010} & \textit{39.7}
  & \textit{.302} & \textit{.246} & \textit{.162} & \textit{.007} & \textit{41.3} \\
\bottomrule
\end{tabular}%
}

\vspace{2pt}

\resizebox{0.67\textwidth}{!}{%
\small
\begin{tabular}{l ccccc | ccccc}
\toprule
& \multicolumn{5}{c|}{\textbf{MAGI-1 24B (I2V)}}
& \multicolumn{5}{c}{\textbf{MAGI-1 24B (V2V)}}\\
\cmidrule(lr){2-6}\cmidrule(lr){7-11}
Verifier
& Sp & SpT & W.Sp & MSE & PhyIQ
& Sp & SpT & W.Sp & MSE & PhyIQ \\
\midrule
\textbf{Baseline}
  & .245 & .155 & .142 & .011 & 29.5
  & .413 & .265 & .288 & .003 & 50.01\\
+ VideoMAE
  & .205 & .122 & .119 & .016 & \hmRl{23.5}
  & .397 & .270 & .276 & .003 & \hmG{52.6}\\
+ Qwen2.5-VL
  & .262 & .134 & .156 & .013 & \hmW{29.7}
  & .401 & .238 & .273 & .003 & \hmG{50.6}\\
+ Qwen3-VL
  & .260 & .174 & .157 & .011 & \hmG{32.3}
  & .415 & .282 & .291 & .003 & \hmG{55.1}\\
+ WMReward~\cite{yuan2026wmreward}
  & .234 & .232 & .148 & .009 & \hmG{34.3}
  & .430 & .314 & .312 & .003 & \hmG{62.29}\\
\best{+ \ours{} (Ours)}
  & \best{.276} & \best{.235} & \best{.181} & \best{.008} & \hmg{\textbf{38.6}}
  & \best{.472} & \best{.336} & \best{.346} & \best{.003} & \best{64.50} \\
\textit{Oracle (Upper bound)}
  & \textit{.306} & \textit{.224} & \textit{.195} & \textit{.009} & \textit{40.2}
  & \textit{.522} & \textit{.363} & \textit{.393} & \textit{.002} & \textit{72.9} \\
\bottomrule
\end{tabular}%
}
\vspace{-8pt}
\end{table*}

\begin{figure}[t]
\centering
\includegraphics[width=0.95\textwidth]{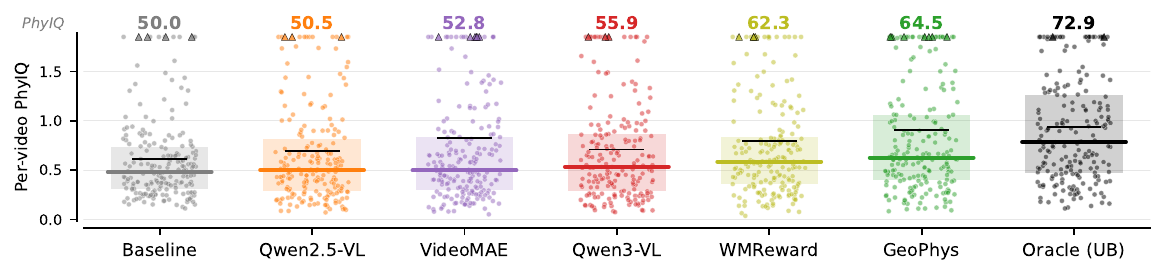}
\caption{\textbf{PhysicsIQ distributions on V2V} (MAGI-1~24B, $N{=}16$).
Each point is one scenario; the coloured bar is the
median, the black the mean, the shaded band the IQR, and triangles are outliers.
\ours{} shifts the whole distribution towards the Oracle ceiling, not only the
mean.}
\label{fig:v2v_distributions}
\end{figure}

\paragraph{Findings.}
\ours{} outperforms every inference-time verifier baseline on all
matched-pool generators (Table~\ref{tab:bon_full}). On MAGI-1~4.5B
it scores $34.0$ ($+15.2$ over no verifier). The next-best verifier,
 Qwen3-VL, scores $30.1$. On Wan2.1~14B, CogVideoX-5B, and
MAGI-1~24B it scores $34.0$, $34.1$, and $38.6$, within $5.7$,
$7.2$, and $1.6$ points of the Oracle upper bound respectively.
No backbone dominates: DINOv3~L18 angle consistency ($+10.32$) and VOneNet~V1 acceleration ($+10.26$) lead on MAGI-1~4.5B, with CORnet-S~IT speed variation ($+7.64$) and DINOv2~L12 angle consistency ($+3.06$) trailing; the OR ensemble reaches $+15.21$ (App.~\ref{app:bon_signals}).
\ours{} also scales
sharply on MAGI-1~24B V2V (Fig.~\ref{fig:v2v_scaling}). As $N$
grows, Qwen2.5-VL and Qwen3-VL plateau near baseline.
WMReward~\cite{yuan2026wmreward} reaches ${\sim}62$ at $N{=}12$.
\ours{} rises to ${\sim}64$. The per-scenario distributions (Fig.~\ref{fig:v2v_distributions}) show the gain is
broad rather than concentrated: \ours{} raises both the median and the upper
quartile above every other real selector and towards the Oracle, so most scenarios
improve, not just a few large wins (\ours{} Qualitative examples in
App.~\ref{app:qualitative}).

\begin{figure}[htb]
\centering
\includegraphics[width=0.9\textwidth]{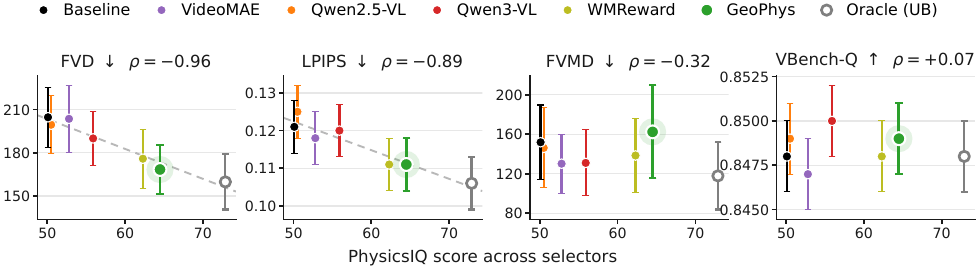}
\caption{
\textbf{Video-quality metrics} (MAGI-1~24B, V2V, $N{=}16$). Each metric
against the PhysicsIQ score; bars are scenario-level
standard errors.
Full $\pm$\,SE values in Table~\ref{tab:metrics_app}.}
\label{fig:metrics_vs_physiq}
\end{figure}

\vspace{-0.5em}
\paragraph{Video-quality metrics.}
We score the V2V MAGI-1~24B ($N{=}16$) on four video-quality
metrics against PhysicsIQ (Fig.~\ref{fig:metrics_vs_physiq}; full $\pm$\,SE in
Table~\ref{tab:metrics_app}), each with a scenario-level standard error (SE):
FVD~\cite{unterthiner2019accurategenerativemodelsvideo} (Fr\'echet distance
between the real and generated clip distributions in I3D feature space),
LPIPS~\cite{zhang2018perceptual} (per-frame perceptual distance to the real
continuation), FVMD~\cite{liu2024frechetvideomotiondistance} (Fr\'echet distance
between tracked-keypoint motion features), and VBench~\cite{huang2023vbench} (a
no-reference quality suite; we report VBench-Q, the mean of its six quality
dimensions). FVD and LPIPS measure fidelity to the real continuation, which a
plausible video matches more closely; both track PhysicsIQ ($\rho=-0.96,-0.89$),
and \ours{} lead on both. FVMD does not track PhysicsIQ
($\rho=-0.32$): its SE of 30 to 47 spans the entire 44-point range, leaving
\ours{} ($162\pm47$) and the baseline ($152\pm38$) indistinguishable. VBench-Q is
flat ($\rho=+0.07$, $\leq0.4\%$ spread within SE), so generic quality is
unchanged; the full breakdown is in Appendix~\ref{app:vbench}. Thus \ours{}'s
gains are confined to the physics-fidelity metrics (FVD, LPIPS); generic quality
(VBench-Q) is unchanged and motion distribution (FVMD) is unresolvable. The only
VBench dimension that moves, dynamic degree, falls as PhysicsIQ rises
($\rho\approx-0.87$), so \ours{} suppresses spurious motion and selects for
physical plausibility, not only visual quality.

\paragraph{Compute footprint.}
\begin{wrapfigure}{r}{0.4\textwidth}
\vspace{-1em}
\centering
\includegraphics[width=\linewidth]{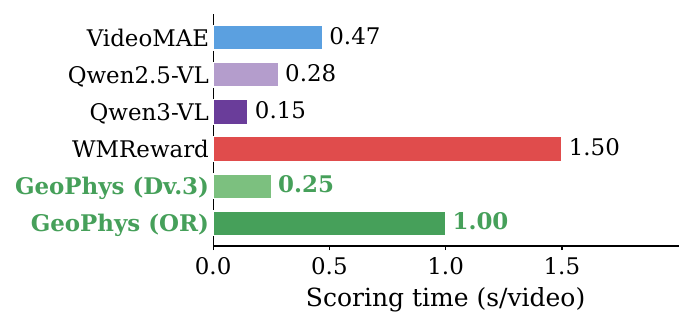}
\caption{\textbf{\ours{} inference time per video}
($N{=}16$, single H100).}
\label{fig:bon_compute}
\vspace{-2em}
\end{wrapfigure}
\ours{}'s verifier path scores at $0.25$\,s/video and $1.2$\,GB VRAM
with a single backbone (DINOv3 ViT-L), and $1.0$\,s/video and
$2.0$\,GB with the ensemble
(Fig.~\ref{fig:bon_compute}; full breakdown in
Appendix~\ref{app:bon_compute}). WMReward takes $1.5$\,s/video and $9.3$\,GB.\ours{} reaches better PhysicsIQ at
$1.5{\times}$ lower wall-clock and $4.65{\times}$ lower memory.
Because test-time scaling compounds the per-candidate cost, this matters: at the same wall-clock budget \ours{} supports roughly $5\times$ more candidates than WMReward, compounding the per-candidate quality gain.

\begin{takeaway}
Test-time compute scales physically plausible video generation, but
only with the right verifier. \ours{} closes the gap to the oracle
ceiling faster than world-model and other verifiers, at $4.65\times$
less compute, using only frozen image encoders.
\end{takeaway}

\vspace{-0.5em}

\section{Discussion and Conclusion}
\vspace{-0.5em}
\label{sec:discussion}
\paragraph{Summary.}
The geometry of feature trajectories through frozen image encoders is
a strong, training-free signal for physical plausibility. A single
score tracks human EEG responses to object-permanence violations
(Sec.~\ref{sec:neuro}); sets SoTA detection on LikePhys ($98.3\%$)
and IntPhys2 ($93.3\%$), surpassing twelve diffusion models, MLLM
judges, and video-pretrained world models
(Sec.~\ref{sec:exp_detection}); and lifts MAGI-1 from 50.01\% to 64.50\% on PhysicsIQ at $1.5\times$ lower wall-clock and
$4.65\times$ lower memory than V-JEPA~2 verifiers
(Sec.~\ref{sec:exp_bon}). For test-time scaling, our results extend
the literature~\cite{snell2024scaling,brown2024large,singhal2025a}
to physically plausible video under a much weaker verifier.
Best-of-$N$ exploits the asymmetry between generation and
verification, reminiscent of the asymmetry that, in its formal limit, 
separates $\mathsf{P}$ from $\mathsf{NP}$: generating a physically
plausible continuation is hard, recognising one need not be. The
fact that a frozen image encoder, orders of magnitude cheaper than
the generator and trained on neither video nor physics, recovers
most of the oracle gap suggests that for many applications the
bottleneck is not sophisticated verification but having any
verifier with the right inductive bias.

\textbf{Implications for neuroscience.}
Peri-occlusion, per-frame \ours{} signals reproduce the time course
and load-scaling of the contralateral delay activity
(CDA)~\cite{balaban2024electrophysiology,liu2024violations}, an EEG
marker of visual working memory. The encoders see only static images,
with no working-memory, object-permanence, or physics objective. One
reading: the CDA reflects IT cortex object-load signatures, which
CORnet-S~IT is modelled after. IT recognises
objects~\cite{dicarlo2012brain}; the CDA scales with object count,
not feature complexity~\cite{balaban2015number}. Geometric features suffice to mimic the CDA's surface properties
without claiming a shared mechanism. 
Replication on larger VOE corpora is the natural next step.

\textbf{Limitations.}
We do not claim frozen backbones represent or simulate physics; \ours{}
is a correlate of plausibility, not an implementation. One shortcoming is that our signals are time-symmetric: a backwards-played video
traces the same feature-space trajectory as the forward original,
consistent with evidence that video encoder features are broadly
time-symmetric~\cite{bagad2026chirality}. Moreover, the test-time-scaling
result is bounded by candidate diversity: PhysicsIQ's oracle ceiling is below
$100\%$, so selection is constrained by what generators produce.
We see \ours{} as a diagnostic for current generators, not a substitute
for genuine world-models~\cite{vafa2025foundationmodelfoundusing,
interno2026observer}.

\section{Acknowledgements}
We would like to thank Andrea Castellani, Sebastian Schmitt, Xavier Bonet-monroig, Linus Ekstrøm, Riccardo Cadei for insightful discussions and feedback. Christian Internò acknowledges funding from the Honda Research Institute Europe. This work was performed with assistance from the US National Institutes of Health Grant S10OD028632-01.

\bibliographystyle{unsrtnat}
\bibliography{references}


\newpage
\appendix
\section*{Appendix Contents}
\vspace{0.5em}
\noindent\rule{\linewidth}{0.8pt}
\vspace{1.2em}

\newcommand{\tocsection}[2]{%
    \noindent\hyperref[#1]{\textbf{\ref*{#1}}\quad#2}\hfill\pageref{#1}\par\vspace{3pt}}
\newcommand{\tocsubsection}[2]{%
    \noindent\hspace{2em}\hyperref[#1]{\ref*{#1}\quad#2}\ \textcolor{black!50}{\dotfill}\ \pageref{#1}\par}

\tocsection{app:straightening}{Perceptual Straightening Analysis}
\tocsection{app:layer_sweep}{Layer Sweep Analysis}
\tocsection{app:add_neuro}{Additional VOE Results}
\tocsection{app:signals}{Signal Breakdown}
\tocsection{app:detection_analysis}{Additional Detection Analysis}
\tocsection{app:physicsiq}{PhysicsIQ Benchmark Protocol}
\tocsection{app:qualitative}{Qualitative Best-of-$N$ Comparisons}
\tocsection{app:bon_signals}{Per-Signal BoN Performance}
\tocsection{app:bon_compute}{BoN Compute Cost}

\vspace{0.3em}
\noindent\rule{\linewidth}{0.4pt}
%
%
%







\section{Perceptual straightening analysis}
\label{app:straightening}

At each backbone's best layer (Table~\ref{tab:detection}), we quantify the plausible-vs-violated difference as a paired Cohen's $d$ together with a paired $t$-test (Table~\ref{tab:app_straightening}). All four backbones produce highly significant differences ($p < 10^{-8}$), with effect sizes ranging from $d = 0.22$ (VOneNet V1) to $d = 0.44$ (CORnet-S~IT). Note that these are \emph{turning-angle} effect sizes (straightening~\cite{henaff2019straightening}).

\begin{table}[h]
\caption{Perceptual-straightening test at each backbone's best layer. $\bar{\phi}^{+}$/$\bar{\phi}^{-}$: mean turning angle on plausible/violated videos. $d$: paired Cohen's $d$. Paired $t$-test over 650 video pairs.}
\label{tab:app_straightening}
\centering
\small
\setlength{\tabcolsep}{5pt}
\begin{tabular}{llccccc}
\toprule
Backbone & Layer & $\bar{\phi}^{+}$ & $\bar{\phi}^{-}$ & $\Delta$ & $d$ & $p$ \\
\midrule
DINOv2 & L12 & 1.742 & 1.807 & $+0.065$ & 0.34 & $2.2\times 10^{-17}$ \\
DINOv3 & L18 & 1.778 & 1.836 & $+0.058$ & 0.34 & $1.7\times 10^{-17}$ \\
CORnet-S & V1 & 1.591 & 1.708 & $+0.117$ & 0.44 & $2.0\times 10^{-26}$ \\
VOneNet & V1 & 2.027 & 2.042 & $+0.016$ & 0.22 & $1.9\times 10^{-\phantom{0}8}$ \\
\bottomrule
\end{tabular}
\end{table}

\begin{figure*}[h]
    \centering
    \includegraphics[width=0.7\textwidth]{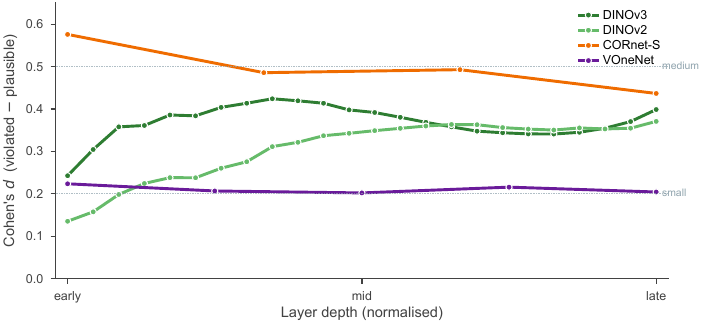}
    \caption{\textbf{Cohen's $d$ of the perceptual-straightening effect across normalised layer depth.} Each point is the paired Cohen's $d$ between turning angles of violated and plausible videos, averaged over 650 LikePhys pairs, at one layer of one backbone. All 57 combinations give $d > 0$ ($p < 10^{-8}$). DINOv2 rises monotonically; CORnet-S peaks at V1 ($d{=}0.58$); DINOv3 peaks mid-depth; VOneNet is flat.}
    \label{fig:app_straightening_depth}
\end{figure*}

The direction of the straightening effect is universal (violated~$>$~plausible, $p<10^{-8}$, all 57 layer and backbone combinations). The magnitude varies with depth and architecture (Fig.~\ref{fig:app_straightening_depth}, Table~\ref{tab:app_straightening_peaks}).

\paragraph{DINOv2~\cite{oquab2024dinov2}}
$d$ rises  from $0.14$ at layer~1 to $0.37$, matching the primate V1$\to$IT profile reported by H\'enaff et al.~\cite{henaff2019straightening,henaff2021straightening}. A self-supervised ViT trained only on multi-view invariance spontaneously acquires the same depth-wise gradient observed in biological vision.

\paragraph{DINOv3~\cite{simoni2025dinov3}}
$d$ peaks at layer~8 ($d = 0.42$) and descends slightly thereafter. The register tokens introduced in DINOv3~\cite{simoni2025dinov3} reorganise late-layer representations to reduce attention artefacts, partially attenuating the geometric-straightness signal.

\paragraph{CORnet-S~\cite{kubilius2019cornet}}
V1 produces the strongest effect of any layer of any backbone ($d = 0.58$), declining  through V2 ($0.49$), V4 ($0.49$), and IT ($0.44$). The layer architecturally designated to mimic primate V1 is also the layer with the sharpest plausibility signal---consistent with the IntPhys2~\cite{bordes2025intphys2} detection results where CORnet-S~V1 is also the strongest CORnet-S layer (61.1\%).

\paragraph{VOneNet~\cite{vonenet2020}}
The fixed Gabor front-end produces a weaker but consistent effect ($d \approx 0.20$), essentially unchanged through its ResNet-50 layers~\cite{jones1987gabor, 7780459}. Because the front-end is not trained, the straightening signal emerges at the level of oriented-edge responses tuned to natural-image statistics and does not require further representational depth.

\begin{table}[h]
\caption{Peak Cohen's $d$ for perceptual straightening per backbone.}
\label{tab:app_straightening_peaks}
\centering
\small
\setlength{\tabcolsep}{5pt}
\begin{tabular}{llccc}
\toprule
Backbone    & Peak layer  & $d$   & $p$                            & Profile \\
\midrule
DINOv2      & L12         & 0.37  & $< 10^{-17}$                   & Monotonic rise \\
DINOv3      & L18          & 0.42  & $< 10^{-21}$                   & Mid-depth peak \\
CORnet-S    & IT          & \textbf{0.58}  & $< 10^{-30}$          & Decline from V1 \\
VOneNet     & V1 & 0.22  & $< 10^{-\phantom{0}8}$         & Flat \\
\bottomrule
\end{tabular}
\end{table}

\section{Layer sweep analysis}
\label{app:layer_sweep}
Backbone weights are taken from
\url{https://github.com/facebookresearch/dinov2} (DINOv2/v3),
\url{https://github.com/dicarlolab/CORnet} (CORnet-S), and
\url{https://github.com/dicarlolab/vonenet} (VOneNet).

\begin{figure*}[h]
    \centering
    \begin{subfigure}[t]{\textwidth}
        \centering
        \includegraphics[width=\textwidth]{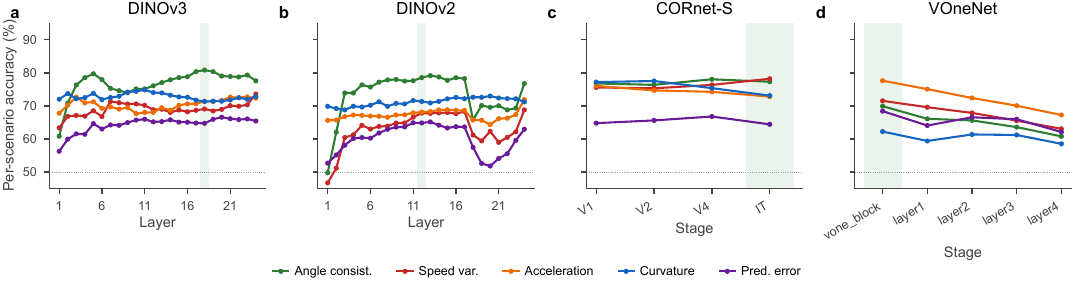}
    \end{subfigure}
    \vspace{0.6em}
    \begin{subfigure}[t]{\textwidth}
        \centering
        \includegraphics[width=\textwidth]{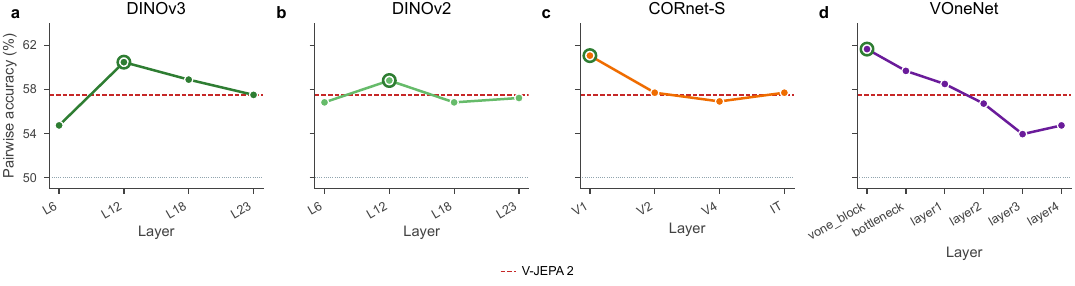}
    \end{subfigure}
    \caption{\textbf{Layer sweep on LikePhys (top) and IntPhys2
    (bottom).} For each backbone, we report per-layer signal accuracy
    and select the readout layer used in the main text. LikePhys
    favours deeper layers for ViTs and IT for CORnet-S; IntPhys2
    favours mid-level (and V1-like) features.}
    \label{fig:app_layer_sweep_combined}
\end{figure*}

\paragraph{LikePhys.}
Fig.~\ref{fig:app_layer_sweep_combined} (top) shows per-scenario accuracy vs.\ layer for each of the five kinematic signals across the four backbones. Two observations motivate the paper's layer choices. First, for the ViT backbones, \emph{angle consistency} peaks at a deep-but-not-final layer (DINOv2~L12, DINOv3~L18) because the very last blocks reorganise features for downstream tasks and lose some of the trajectory regularity. Second, for CORnet-S, \emph{speed variation} grows with depth and peaks at IT, consistent with the recurrent module building temporal sensitivity. VOneNet's best signal is acceleration at the fixed Gabor front-end (V1), which picks up on local curvature changes directly.

\paragraph{IntPhys2.}
Fig.~\ref{fig:app_layer_sweep_combined} (bottom) shows per-layer
best-signal accuracy on IntPhys2, where the best signal at each layer
is the strongest of the five temporal signals in $\Phi_{\mathrm{temp}}$
(Eq.~\ref{eq:temporal_descriptor}). The trend is the opposite of
LikePhys: \emph{mid-level features outperform late features}. DINOv3
L12 (60.5\%) beats L18 (58.9\%) and L23 (57.5\%); CORnet-S V1
(61.1\%) beats IT (57.7\%). Localised semantic violations---an
object vanishing behind an occluder, a ball passing through a
wall---are best captured by the mid-level features of DINOv2/v3 and
by the V1-like layers of CORnet-S and VOneNet, where local
trajectory structure is most cleanly preserved.

\section{Additional VOE results}
\label{app:add_neuro}

In Fig.~\ref{fig:4_Neuro}, we demonstrate model-brain alignment on object permanence 
violations in which one object enters occlusion but two exit (the `Create' scenario), and matched control videos. 
In Fig.~\ref{fig12: neuro_vanish_ts}, we present a complementary scenario, `Vanish,' 
in which two objects enter occlusion but only one exits. For both scenarios, we report 
the average difference between \ours{} signals on valid and invalid videos post-occlusion in 
Table~\ref{tab:prepost_transposed}, baseline corrected so that the pre-occlusion 
difference (when videos are approximately identical) is exactly zero. This controls 
for minor variations introduced by ADEPT rendering. Finally, 
Table~\ref{tab:heatmap_neuro} reports overall \ours{} detection performance on 
additional VOE datasets from neuroscience.

\begin{center}
\includegraphics[width=1\textwidth]{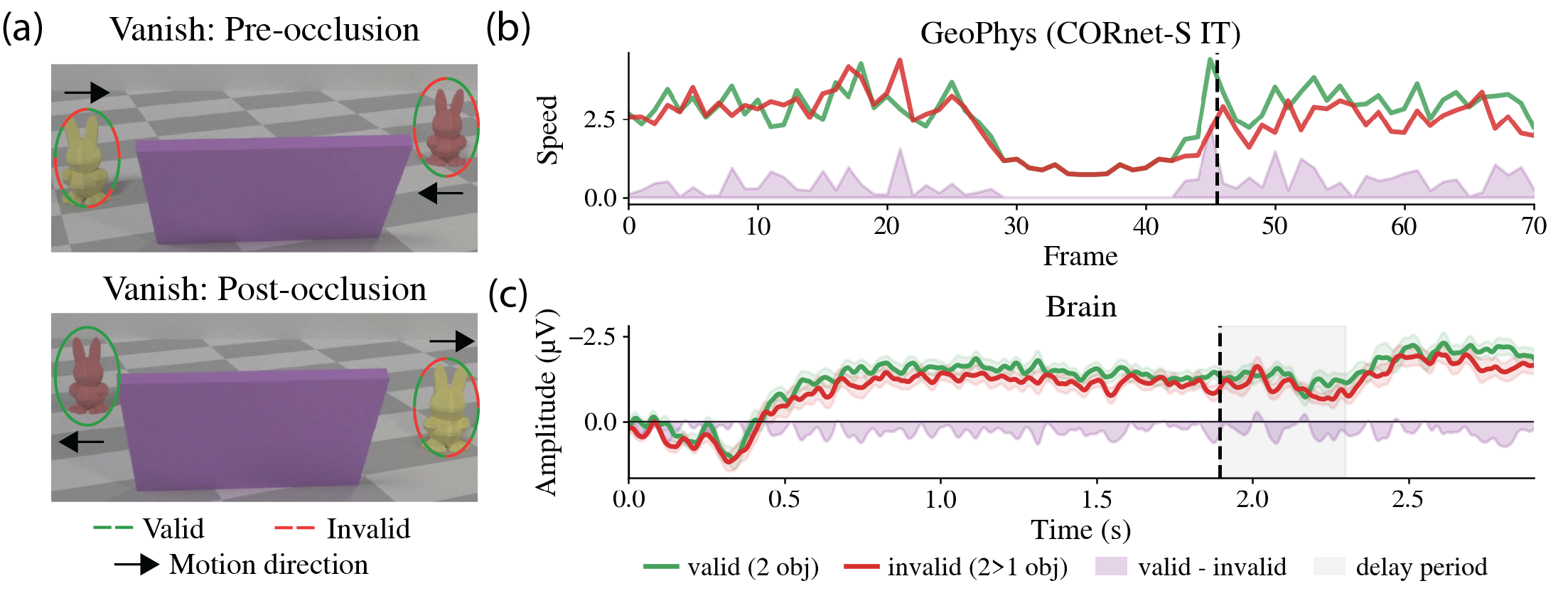}
\end{center}
\captionof{figure}{
\textbf{Detecting object permanence violations in models and brains: Vanish condition }
(a) Example VOE stimuli for the Vanish scenario: in valid videos, two objects enter and exit occlusion; in the invalid videos, two objects enter but one exits. (b–c) valid and invalid Vanish signals from \ours{} CORnet-S IT (speed) (b) and EEG contralateral delay activity (from \cite{balaban2024electrophysiology}; c).  Both \ours{} and brain signals are elevated for the valid condition after occlusion offset (dashed line).}
\label{fig12: neuro_vanish_ts}

\begin{table}[h]
\centering
\scriptsize
\renewcommand{\arraystretch}{1.3}
\setlength{\tabcolsep}{1pt}
\begin{tabular}{l*{5}{r}@{\hspace{3pt}}*{5}{r}@{\hspace{3pt}}*{5}{r}@{\hspace{3pt}}*{5}{r}}
\toprule
 & \multicolumn{5}{c}{\textit{CORnet-S IT}} & \multicolumn{5}{c}{\textit{DINOv2}} & \multicolumn{5}{c}{\textit{DINOv3}} & \multicolumn{5}{c}{\textit{VOneNet}} \\
\cmidrule(lr){2-6}\cmidrule(lr){7-11}\cmidrule(lr){12-16}\cmidrule(lr){17-21}
 & \rotatebox{45}{\tiny Spd} & \rotatebox{45}{\tiny Acc} & \rotatebox{45}{\tiny Curv} & \rotatebox{45}{\tiny Ang} & \rotatebox{45}{\tiny Pred}
 & \rotatebox{45}{\tiny Spd} & \rotatebox{45}{\tiny Acc} & \rotatebox{45}{\tiny Curv} & \rotatebox{45}{\tiny Ang} & \rotatebox{45}{\tiny Pred}
 & \rotatebox{45}{\tiny Spd} & \rotatebox{45}{\tiny Acc} & \rotatebox{45}{\tiny Curv} & \rotatebox{45}{\tiny Ang} & \rotatebox{45}{\tiny Pred}
 & \rotatebox{45}{\tiny Spd} & \rotatebox{45}{\tiny Acc} & \rotatebox{45}{\tiny Curv} & \rotatebox{45}{\tiny Ang} & \rotatebox{45}{\tiny Pred} \\
\midrule
Create $\bar{x}$ & -0.25 & -0.33 & -0.03 & +0.03 & -0.33 & -0.00 & -0.01 & -0.03 & +0.03 & -0.01 & +0.13 & +0.29 & +0.02 & -0.02 & +0.29 & +0.00 & +0.00 & -0.01 & +0.01 & +0.00 \\
Create $\sigma$  & 0.13 & 0.32 & 0.09 & 0.09 & 0.32 & 0.01 & 0.01 & 0.08 & 0.07 & 0.01 & 0.17 & 0.50 & 0.14 & 0.13 & 0.50 & 0.01 & 0.02 & 0.03 & 0.03 & 0.02 \\
\midrule
Vanish $\bar{x}$ & +0.16 & +0.16 & -0.01 & +0.01 & +0.16 & +0.00 & +0.01 & +0.06 & -0.06 & +0.01 & -0.04 & +0.01 & +0.06 & -0.06 & +0.01 & +0.00 & +0.00 & -0.02 & +0.02 & +0.00 \\
Vanish $\sigma$  & 0.14 & 0.21 & 0.05 & 0.05 & 0.21 & 0.01 & 0.03 & 0.10 & 0.10 & 0.03 & 0.15 & 0.28 & 0.06 & 0.06 & 0.28 & 0.01 & 0.02 & 0.03 & 0.03 & 0.02 \\
\bottomrule
\end{tabular}
\vspace{4pt}
\caption{Mean valid $-$ invalid \ours{} signal post-occlusion (baseline-subtracted).
$\bar{x}$/$\sigma$ across 7 pairs.
Spd=Speed, Acc=Acceleration, Curv=Curvature, Ang=Angle Cons., Pred=Pred.\ Error.}
\label{tab:prepost_transposed}
\end{table}


\newcommand{\hmcellfrac}[3]{%
  \ifnum#3=100\cellcolor[HTML]{9bc398}\textbf{#1/#2}%
  \else\ifnum#3>82\cellcolor[HTML]{b9d6b7}#1/#2%
  \else\ifnum#3>66\cellcolor[HTML]{d7e9d6}#1/#2%
  \else\ifnum#3>57\cellcolor[HTML]{eef6ee}#1/#2%
  \else\ifnum#3=50\cellcolor[HTML]{ffffff}#1/#2%
  \else\ifnum#3>44\cellcolor[HTML]{ffffff}#1/#2%
  \else\ifnum#3>32\cellcolor[HTML]{f4c8c8}#1/#2%
  \else\cellcolor[HTML]{e87878}#1/#2%
  \fi\fi\fi\fi\fi\fi\fi%
}

\begin{table}[h]
\centering
\footnotesize
\renewcommand{\arraystretch}{1.4}
\setlength{\tabcolsep}{2pt}
\caption{\textbf{Neural signal alignment detection.}
Pairwise accuracy ($n_\text{correct}/n_\text{total}$, $\uparrow$ 0\%
\protect\tikz[baseline=-0.5ex]{
  \protect\shade[left color=hm19, right color=hm60] (0,0) rectangle (0.75,0.2);
  \protect\shade[left color=hm60, right color=hm100] (0.75,0) rectangle (1.5,0.2);
} 100\%) on VOE datasets. Perman.\ = Permanence, Solid.\ = Solidity, Mat.\ Viol.\ = Material Violations.}
\begin{tabular}{l ccc c}
\toprule
& \textbf{Balaban}~\cite{balaban2024electrophysiology} & \multicolumn{2}{c}{\textbf{Liu}~\cite{liu2024violations}} & \textbf{Kaiser}~\cite{kaiser2023eeg} \\
\cmidrule(lr){2-2}\cmidrule(lr){3-4}\cmidrule(lr){5-5}
& Perman. & Perman. & Solid. & Mat.\ Viol. \\
\midrule
\multicolumn{5}{l}{\textit{GeoPhys individual}} \\
DINOv2   & \hmcellfrac{2}{2}{100} & \hmcellfrac{4}{6}{67} & \hmcellfrac{3}{6}{50} & \hmcellfrac{2}{4}{50} \\
DINOv3   & \hmcellfrac{1}{2}{50}  & \hmcellfrac{5}{6}{83} & \hmcellfrac{4}{6}{67} & \hmcellfrac{0}{4}{0}  \\
CORnet-S & \hmcellfrac{1}{2}{50}  & \hmcellfrac{2}{6}{33} & \hmcellfrac{4}{6}{67} & \hmcellfrac{2}{4}{50} \\
VOneNet  & \hmcellfrac{2}{2}{100} & \hmcellfrac{3}{6}{50} & \hmcellfrac{3}{6}{50} & \hmcellfrac{3}{4}{75} \\
\midrule
\multicolumn{5}{l}{\textit{GeoPhys ensembles}} \\
Majority & \hmcellfrac{1}{2}{50}  & \hmcellfrac{3}{6}{50}  & \hmcellfrac{4}{6}{67}  & \hmcellfrac{2}{4}{50}  \\
OR       & \hmcellfrac{2}{2}{100} & \hmcellfrac{6}{6}{100} & \hmcellfrac{6}{6}{100} & \hmcellfrac{4}{4}{100} \\
\bottomrule
\end{tabular}
\label{tab:heatmap_neuro}
\end{table}

\section{Signal breakdown}
\label{app:signals}
Table~\ref{tab:signals} reports the full accuracy and effect size for
every backbone $\times$ \ours{} signal combination on LikePhys.
Each backbone specialises in a different signal: CORnet-S on speed
variation ($d{=}0.96$, the largest single-signal effect size),
DINOv2/v3 on angle consistency ($d{=}0.73$--$0.76$), and VOneNet on
acceleration ($d{=}0.37$). Prediction error provides moderate but
consistent signal across all backbones ($d{=}0.25$--$0.42$), never
the best but never below chance.

\begin{table*}[h]
\caption{Signal breakdown on \textbf{LikePhys}: accuracy~(\%) and
Cohen's~$|d|$ for every backbone $\times$ kinematic signal.
\best{Green}: best signal per backbone.}
\label{tab:signals}
\centering
\small
\setlength{\tabcolsep}{3.5pt}
\begin{tabular}{l cc cc cc cc cc}
\toprule
& \multicolumn{2}{c}{Curvature}
& \multicolumn{2}{c}{Speed var.}
& \multicolumn{2}{c}{Acceleration}
& \multicolumn{2}{c}{Angle consist.}
& \multicolumn{2}{c}{Pred.\ error} \\
\cmidrule(lr){2-3} \cmidrule(lr){4-5} \cmidrule(lr){6-7} \cmidrule(lr){8-9} \cmidrule(lr){10-11}
& Acc & $|d|$
& Acc & $|d|$
& Acc & $|d|$
& Acc & $|d|$
& Acc & $|d|$ \\
\midrule
DINOv2 L12
  & 71 & .40  & 68 & .43  & 68 & .31
  & \best{79} & \best{.73}  & 66 & .30 \\
DINOv3 L18
  & 71 & .41  & 69 & .55  & 71 & .34
  & \best{81} & \best{.76}  & 67 & .32 \\
CORnet-S IT
  & 73 & .51  & \best{78} & \best{.96}  & 73 & .58
  & 77 & .68  & 70 & .42 \\
VOneNet V1
  & 62 & .13  & 72 & .46  & \best{78} & \best{.37}
  & 70 & .36  & 64 & .25 \\
\bottomrule
\end{tabular}
\end{table*}



\section{Additional detection analysis}
\label{app:detection_analysis}

\paragraph{LikePhys: per-physics-domain accuracy.}
Table~\ref{tab:per_domain} groups the 12 LikePhys scenarios by physical domain following~\cite{yuan2025likephys}. Each backbone specialises: VOneNet dominates rigid-body and optical domains (92.4\%, 74.5\%), DINOv3 dominates fluid (88.0\%), and DINOv2/CORnet-S share continuum (86.5\%/86.0\%). The majority vote recovers 80--96\% across all four domains, compared to 43.5--63.6\% for the best VDM baseline (Hunyuan T2V).

\begin{table}[h]
\caption{\textbf{LikePhys: per-physics-domain accuracy (\%).}
\underline{Underline}: best VDM baseline (Hunyuan T2V).
\textbf{Bold}: best \ours{} backbone per domain.}
\label{tab:per_domain}
\centering
\small
\setlength{\tabcolsep}{4pt}
\begin{tabular}{l cccc c}
\toprule
 & Rigid & Contin. & Fluid & Optical & Overall \\
\midrule
Hunyuan T2V & \underline{63.6} & \underline{43.5} & \underline{51.3} & \underline{59.0} & \underline{56.4} \\
\midrule
DINOv2 L12   & 86.8 & \textbf{86.5} & 71.3 & 61.0 & 78.6 \\
DINOv3 L18   & 85.0 & \textbf{86.5} & \textbf{88.0} & 54.0 & 80.8 \\
CORnet-S IT  & 89.8 & 86.0 & 65.0 & 61.0 & 78.2 \\
VOneNet V1   & \textbf{92.4} & 74.5 & 57.3 & \textbf{74.5} & 77.6 \\
\midrule
Majority     & 94.6 & 95.5 & 92.7 & 80.5 & 90.9 \\
\bottomrule
\end{tabular}
\end{table}

\paragraph{IntPhys2: accuracy by difficulty and camera type.}
Table~\ref{tab:intphys2_splits} breaks down IntPhys2 results by the Easy/Medium/Hard sub-splits and Fixed/Moving camera configurations defined in~\cite{bordes2025intphys2}. Unlike VLM baselines that degrade from Easy to Hard (Gemini-2.5~Flash: 64.4\%$\to$54.5\%), \ours{}'s majority vote is stable across difficulty levels (69--70\%). Fixed-camera scenes are slightly easier (71.4\% vs.\ 67.7\%), particularly for CORnet-S~V1, whose retinotopic structure benefits from spatial stability.

\begin{table}[h]
\caption{\textbf{IntPhys2: accuracy (\%) by difficulty and camera type.}
Published baselines from~\cite{bordes2025intphys2}.}
\label{tab:intphys2_splits}
\centering
\small
\setlength{\tabcolsep}{3pt}
\begin{tabular}{l ccc cc}
\toprule
& \multicolumn{3}{c}{By difficulty} & \multicolumn{2}{c}{By camera} \\
\cmidrule(lr){2-4} \cmidrule(lr){5-6}
& Easy & Med. & Hard & Fixed & Moving \\
\midrule
\multicolumn{6}{l}{\textit{Published baselines}} \\
GPT-4o          & 57.7 & 54.8 & 54.2 & 57.2 & 57.7 \\
Gemini-2.5 Fl.  & \underline{64.4} & 56.8 & 54.5 & 58.7 & 58.6 \\
V-JEPA 2        & 54.0 & \underline{58.5} & \underline{59.4} & 54.8 & \underline{62.4} \\
\midrule
\multicolumn{6}{l}{\textit{\ours{} individual}} \\
DINOv2 L12    & 50.0 & 56.2 & 52.7 & 53.1 & 56.2 \\
DINOv3 L12    & 51.9 & 53.8 & 53.6 & 55.8 & 51.2 \\
CORnet-S V1   & 55.8 & \textbf{62.0} & 59.8 & \textbf{62.7} & 54.5 \\
VOneNet V1    & 52.9 & 52.0 & 50.0 & 50.5 & 50.0 \\
\midrule
\multicolumn{6}{l}{\textit{\ours{} ensembles}} \\
Majority      & 69.6 & 69.5 & 68.8 & 71.4 & 67.7 \\
OR            & 85.3 & 92.0 & 89.9 & 91.3 & 91.1 \\
\midrule
Human         & 96.2 & 97.8 & 95.5 & -- & -- \\
\bottomrule
\end{tabular}
\end{table}

\section{PhysicsIQ benchmark protocol}
\label{app:physicsiq}

This appendix details the PhysicsIQ benchmark~\cite{motamed2025physicsiq}
used in Sec.~\ref{sec:exp_bon} to evaluate \ours{} as a best-of-$N$
verifier for video generation.

\paragraph{Dataset.}
PhysicsIQ contains 396 high-quality videos covering 66 real-world
physical scenarios across five categories: solid mechanics (38
scenarios), fluid dynamics (15), optics (8), thermodynamics (3), and
magnetism (2). Each scenario was filmed from three perspectives
(left, center, right) on a static Sony Alpha a6400 with a 16--50\,mm
lens, at $30$\,FPS and $3840 \times 2160$ resolution. Each scenario
was shot \emph{twice} under identical conditions to capture the
inherent randomness of real-world physical interactions. The benchmark
spans $198$ evaluation scenarios ($66 \times 3$ perspectives), each
$8$ seconds long.

\paragraph{Conditioning protocol.}
Each $8$-second video is split into a $3$-second conditioning window
and a held-out $5$-second test continuation that serves as ground
truth. Image-to-video (I2V) generators receive only the last frame
of the conditioning window (the \emph{switch frame}) which is
hand-selected per scenario such that the physical event is set up but has not yet occurred (e.g., the first domino is tipped but has not contacted the second). Video-to-video (V2V) generators receive
the full $3$-second clip as multi-frame conditioning. Generators
that accept text receive a human-written description of the
conditioning frames that does not give away the continuation;
generators that do not (e.g., Stable Video Diffusion) receive only
the visual conditioning.

\paragraph{Motion-mask metrics.}
The benchmark quantifies physical realism via four motion-mask
metrics computed between the generated continuation and the real
continuation. A binary $h \times w \times t$ \emph{motion-mask
video} is first extracted by thresholding pixel intensity changes
across frames; the four metrics summarise this mask in different
ways:
\begin{itemize}
\item \textbf{Spatial IoU} (\emph{where} action happens). Collapse
the binary motion mask across time via a max, giving an $h \times
w$ binary motion map. Compute IoU against the real motion map:
\begin{equation}
\text{Spatial-IoU} =
\frac{|M^{\text{sp}}_{\text{real}} \cap M^{\text{sp}}_{\text{gen}}|}
     {|M^{\text{sp}}_{\text{real}} \cup M^{\text{sp}}_{\text{gen}}|}.
\end{equation}

\item \textbf{Spatiotemporal IoU} (\emph{where and when} action
happens). Frame-by-frame IoU on the $h \times w \times t$ binary
mask, averaged across $t$. A model that gets the location right but
the timing wrong scores well on Spatial IoU but poorly here.
\begin{equation}
\text{ST-IoU} =
\frac{|M_{\text{real}} \cap M_{\text{gen}}|}
     {|M_{\text{real}} \cup M_{\text{gen}}|}.
\end{equation}

\item \textbf{Weighted Spatial IoU} (\emph{where and how much}
action happens). Same as Spatial IoU, but collapse time using the
per-frame action density. The metric is the pixel-wise minimum over
maximum, distinguishing repeated motion (e.g., pendulum) from
one-pass motion (e.g., rolling ball):
\begin{equation}
\text{W-IoU} =
\frac{\sum_i \min\!\bigl(M^{\text{w}}_{\text{real},i},\,
                         M^{\text{w}}_{\text{gen},i}\bigr)}
     {\sum_i \max\!\bigl(M^{\text{w}}_{\text{real},i},\,
                         M^{\text{w}}_{\text{gen},i}\bigr)}.
\end{equation}

\item \textbf{MSE} (\emph{how} action happens). Pixel-level mean
squared error between generated and real frames. Strict appearance
similarity. Sensitive to colour and texture hallucinations. Lower
is better:
\begin{equation}
\text{MSE} =
\frac{1}{n} \sum_{i=1}^{n} \bigl(f_{\text{real},i} -
                                f_{\text{gen},i}\bigr)^2.
\end{equation}
\end{itemize}

\paragraph{Aggregate score.}
The four metrics are combined into a single \emph{PhysicsIQ score}
by summing the three IoU metrics and subtracting MSE (with a sign
flip since MSE is inverted), then normalising so that the
\emph{physical variance} scores $100\%$. This
defines an empirical ceiling: a model achieving $100\%$ would be
indistinguishable in motion-mask terms from a real second take. A
score of $0\%$ corresponds to no overlap with reality.

\section{Qualitative best-of-$N$ comparisons}
\label{app:qualitative}
\begin{figure}[p]
\centering
\includegraphics[width=\textwidth]{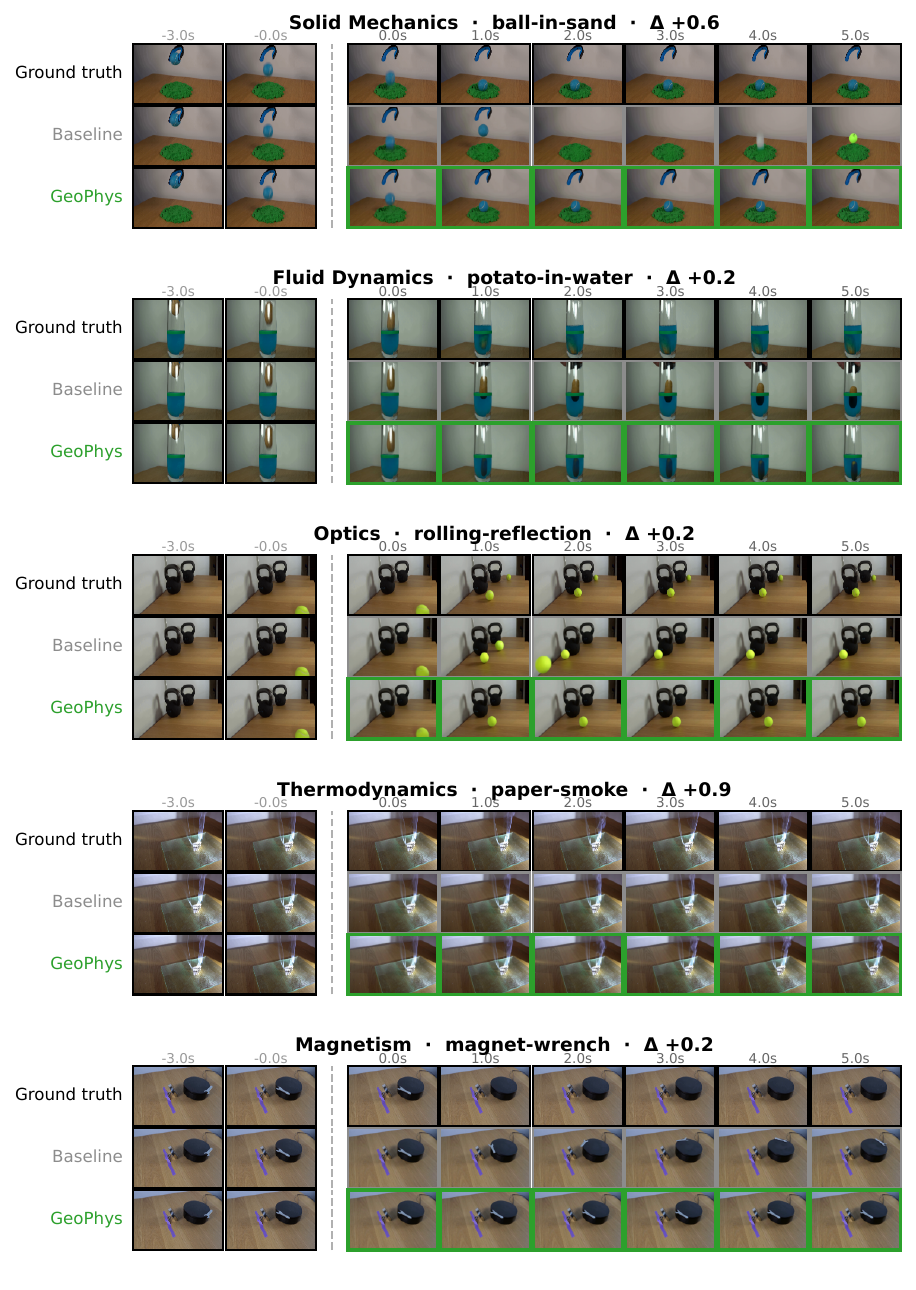}
\caption{\textbf{Qualitative best-of-$N$ selection across physics families}
(PhysicsIQ V2V, MAGI-1~24B, $N{=}16$). Columns left of the dashed line are the shared real conditioning;
columns to its right are the continuations, with the real take
(\emph{Ground truth}) for reference.}
\label{fig:qual_grid}
\end{figure}

Beyond the aggregate scores of Sec.~\ref{sec:exp_bon}, we show what the \ours{}
selection looks like for each of the five PhysicsIQ categories (solid mechanics,
fluid dynamics, optics, thermodynamics, magnetism) beside the ground truth
(Fig.~\ref{fig:qual_grid}). Columns left of the dashed line are the shared
$3$\,s real conditioning; columns to its right are the $5$\,s continuation,
sampled evenly; the per-panel $\Delta$ gives the PhysicsIQ gain in points.

\section{Per-signal \ours{} BoN performance}
\label{app:bon_signals}
We report the per-backbone breakdown of \ours{} as a
best-of-$N$ verifier on the three matched-comparison generators of
Table~\ref{tab:bon_full}. 

\paragraph{Each backbone helps on a different scenario subset.}
On every generator, every individual backbone already beats the
no-verifier baseline by $+5$--$+11$ PhysicsIQ points
(Table~\ref{tab:bon_signals}), but no single backbone dominates
across all 198 scenarios. The OR ensemble lifts performance by an
additional $+4$--$+5$ points over the strongest single backbone,
mirroring the complementarity pattern observed for detection
(Fig.~\ref{fig:complementarity}): different backbones flag different
violations, and selecting the most-confident backbone per scenario
recovers what each misses.

\paragraph{Single-signal gains.}
On every generator, the strongest single backbone is the one whose
detection-winning signal is second-order or near-second-order:
DINOv3 L18 angle consistency (which depends on consecutive
displacements and is therefore second-order in the trajectory) and
VOneNet V1 acceleration each deliver $+8$ to $+10$ PhysicsIQ points,
and CORnet-S IT speed variation contributes a comparable $+7$ to
$+11$.

\begin{table*}[h]
\caption{\textbf{Per-backbone \ours{} BoN performance on the matched
candidate pools.} Each row applies the detection-winning
backbone--signal pair from Table~\ref{tab:signals} as the BoN verifier
on the same $N{=}16$ candidate pool as in
Table~\ref{tab:bon_full}; no signal-search, no PhysicsIQ-specific
tuning. The Ensemble row is the OR rule of
Sec.~\ref{sec:exp_detection} applied across the four backbones.
$\Delta$: gain over the matched no-verifier baseline.}
\label{tab:bon_signals}
\centering
\setlength{\tabcolsep}{4pt}
\renewcommand{\arraystretch}{0.95}
\resizebox{\textwidth}{!}{%
\footnotesize
\begin{tabular}{l l ccccc c}
\toprule
\multirow{2}{*}{Generator} & \multirow{2}{*}{Backbone (signal)}
  & \multicolumn{3}{c}{IoU $\uparrow$}
  & \multirow{2}{*}{MSE $\downarrow$}
  & \multirow{2}{*}{Score $\uparrow$}
  & \multirow{2}{*}{$\Delta$} \\
\cmidrule(lr){3-5}
& & Spatial & Spatiotemp. & Weighted Sp. & & & \\
\midrule
\multirow{6}{*}{MAGI-1 4.5B}
  & Baseline (no verifier)                              & 0.143 & 0.133 & 0.070 & 0.019 & 18.75 & --- \\
  & DINOv2 L12 (angle consist.)                       & 0.135 & 0.178 & 0.069 & 0.011 & 21.81 & +3.06 \\
  & DINOv3 L18 (angle consist.)                       & 0.130 & 0.281 & 0.082 & 0.008 & 29.07 & +10.32 \\
  & CORnet-S IT (speed var.)                          & 0.114 & 0.266 & 0.068 & 0.008 & 26.39 & +7.64 \\
  & VOneNet V1 (accel.)                               & 0.132 & 0.277 & 0.084 & 0.008 & 29.01 & +10.26 \\
  & \best{Ensemble (OR)}                              & \best{0.225} & \best{0.232} & \best{0.128} & \best{0.007} & \best{33.96} & \best{+15.21} \\
\cmidrule(lr){2-8}
\multirow{6}{*}{Wan2.1 14B}
  & Baseline (no verifier)                              & 0.143 & 0.133 & 0.070 & 0.018 & 18.85 & --- \\
  & DINOv2 L12 (angle consist.)                       & 0.135 & 0.178 & 0.069 & 0.010 & 21.95 & +3.10 \\
  & DINOv3 L18 (angle consist.)                       & 0.130 & 0.281 & 0.082 & 0.007 & 29.11 & +10.26 \\
  & CORnet-S IT (speed var.)                          & 0.185 & 0.215 & 0.112 & 0.008 & 29.81 & +10.96 \\
  & VOneNet V1 (accel.)                               & 0.223 & 0.136 & 0.119 & 0.009 & 27.33 & +8.48 \\
  & \best{Ensemble (OR)}                              & \best{0.225} & \best{0.232} & \best{0.128} & \best{0.007} & \best{34.01} & \best{+15.16} \\
\cmidrule(lr){2-8}
\multirow{6}{*}{CogVideoX-5B}
  & Baseline (no verifier)                              & 0.143 & 0.133 & 0.070 & 0.008 & 19.86 & --- \\
  & DINOv2 L12 (angle consist.)                       & 0.135 & 0.178 & 0.069 & 0.008 & 22.04 & +2.18 \\
  & DINOv3 L18 (angle consist.)                       & 0.130 & 0.281 & 0.082 & 0.007 & 29.15 & +9.29 \\
  & CORnet-S IT (speed var.)                          & 0.185 & 0.215 & 0.112 & 0.007 & 29.87 & +10.01 \\
  & VOneNet V1 (accel.)                               & 0.223 & 0.136 & 0.119 & 0.007 & 27.57 & +7.71 \\
  & \best{Ensemble (OR)}                              & \best{0.225} & \best{0.232} & \best{0.128} & \best{0.006} & \best{34.08} & \best{+14.22} \\
\bottomrule
\end{tabular}%
}
\end{table*}
\subsection{Full video-quality metrics}
\label{app:metrics}
Table~\ref{tab:metrics_app} gives the complete numbers behind
Fig.~\ref{fig:metrics_vs_physiq}, with a scenario-level standard error on every
cell. For the Frechet metrics (FVD, FVMD) we resample the 198 scenarios with
replacement and recompute the distance on each resample; for the per-scenario
means (LPIPS, VBench-Q) the standard error is $\mathrm{std}/\sqrt{198}$.
PhysicsIQ is the reference axis and is reported as a point estimate. The
two-family split of the main text is visible directly in the intervals. FVD and
LPIPS have standard errors well below the between-selector spread, so their
ordering against PhysicsIQ is stable: \ours{} attains the lowest FVD and the
second-lowest LPIPS among the learnable selectors. FVMD instead has standard
errors of 30 to 47, comparable to its full 44-point range, so the selectors
overlap and the metric cannot rank them; \ours{} at $162.4\pm47.0$ and the
baseline at $151.9\pm37.7$ are statistically indistinguishable. VBench-Q varies
by 0.4\% across selectors, within its $\pm0.002$ standard error. The Spearman
$\rho$ row is taken across the seven selectors; with $n=7$ these correlations
indicate the trend rather than a precise estimate, and the per-cell intervals
carry the argument.

\begin{table}[h]
\caption{\textbf{Video-quality metrics on PhysicsIQ V2V with scenario-level
$\pm$\,SE} (MAGI-1~24B, $N{=}16$). Point $\pm$ standard error over the 198
scenarios; scenario bootstrap for the Frechet metrics (FVD, FVMD) and
$\mathrm{std}/\sqrt{n}$ for the means (LPIPS, VBench-Q). VBench-Q is the mean of
VBench's six quality dimensions (breakdown in Table~\ref{tab:vbench}). Spearman
$\rho$ is vs.\ PhysicsIQ across selectors.}
\label{tab:metrics_app}
\centering\small
\setlength{\tabcolsep}{6pt}\renewcommand{\arraystretch}{1.2}
\begin{tabular}{@{}l c cccc@{}}
\toprule
Selector & PhyIQ\,$\uparrow$ & FVD\,$\downarrow$ & FVMD\,$\downarrow$ & LPIPS\,$\downarrow$ & VBench-Q\,$\uparrow$ \\
\midrule
Baseline   & 50.1 & $204.7\pm20.9$ & $151.9\pm37.7$ & $0.121\pm0.007$ & $0.848\pm0.002$ \\
VideoMAE   & 52.8 & $203.6\pm23.2$ & $130.2\pm29.9$ & $0.118\pm0.007$ & $0.847\pm0.002$ \\
Qwen2.5-VL & 50.5 & $199.5\pm20.2$ & $146.2\pm40.5$ & $0.125\pm0.007$ & $0.849\pm0.002$ \\
Qwen3-VL   & 55.9 & $190.0\pm18.6$ & $131.0\pm33.4$ & $0.120\pm0.007$ & $0.850\pm0.002$ \\
WMReward   & 62.3 & $173.9\pm20.5$ & $138.5\pm37.2$ & $\mathbf{0.111\pm0.007}$ & $0.848\pm0.002$ \\
\ours{}    & \textbf{64.5} & $\mathbf{168.4\pm17.3}$ & $162.4\pm47.0$ & $\mathbf{0.111\pm0.007}$ & $0.849\pm0.002$ \\
\midrule
\textit{Oracle (UB)} & \textit{72.9} & \textit{$159.8\pm19.3$} & \textit{$117.9\pm34.2$} & \textit{$0.106\pm0.007$} & \textit{$0.848\pm0.002$} \\
\midrule
Spearman $\rho$ & --- & $-0.96$ & $-0.32$ & $-0.89$ & $+0.07$ \\
\bottomrule
\end{tabular}
\end{table}

\subsection{VBench dimension breakdown}
\label{app:vbench}
\begin{table}[h]
\caption{\textbf{VBench dimensions on PhysicsIQ V2V} (MAGI-1~24B, $N{=}16$),
per-dimension mean over the 198 scenarios; the typical scenario-level standard
error is in each row label. Six dimensions are saturated and effectively
constant across selectors; only dynamic degree varies.}
\label{tab:vbench}
\centering\small\setlength{\tabcolsep}{4pt}\renewcommand{\arraystretch}{1.15}
\resizebox{\linewidth}{!}{%
\begin{tabular}{l ccccccc}
\toprule
Dimension & Baseline & VideoMAE & Qwen2.5 & Qwen3 & WMReward & \ours{} & Oracle \\
\midrule
\textit{PhysicsIQ} ($\uparrow$, ref.) & \textit{50.1} & \textit{52.8} & \textit{50.5} & \textit{55.9} & \textit{62.3} & \textit{64.5} & \textit{72.9} \\
\midrule
Subject consist. ($\pm.003$)    & .959 & .956 & .959 & .960 & .959 & .960 & .957 \\
Background consist. ($\pm.001$) & .972 & .971 & .973 & .973 & .972 & .973 & .971 \\
Motion smooth. ($\pm.0001$)     & .9970 & .9970 & .9970 & .9971 & .9971 & .9972 & .9970 \\
Aesthetic ($\pm.006$)           & .472 & .468 & .476 & .476 & .469 & .475 & .469 \\
Imaging /100 ($\pm.5$)          & 68.9 & 68.9 & 69.1 & 69.4 & 69.2 & 69.2 & 69.3 \\
Temporal flicker ($\pm.0001$)   & .9981 & .9982 & .9981 & .9982 & .9985 & .9984 & .9983 \\
\midrule
Dynamic degree ($\pm.018$)      & .096 & .071 & .076 & .071 & .040 & .040 & .066 \\
\bottomrule
\end{tabular}}
\end{table}
Six dimensions are saturated near their ceilings and vary by at most a few
tenths of a percent across selectors, so they cannot distinguish the selection
strategies. Three of these, motion smoothness, temporal flickering and imaging
quality, show a high rank correlation with PhysicsIQ ($\rho>0.7$), but this is an
artefact of ranking values identical to three or four decimal places; their
between-selector differences are within one to two standard errors, so the
correlation orders essentially equal numbers rather than reflecting a real
effect. The only dimension with substantial variation is dynamic degree, which
ranges over a factor of two and anti-correlates with PhysicsIQ ($\rho=-0.87$):
the more physically plausible selections contain less spurious motion. This
matches the FVMD result, where the same suppression of excess motion reads as a
divergence from the ground-truth motion distribution. We therefore summarise
VBench in the main table by VBench-Q, the mean of the six quality dimensions,
and report dynamic degree here, since it measures motion volume rather than
perceptual quality.

\section{BoN compute cost (full)}
\label{app:bon_compute}

Table~\ref{tab:bon_compute_app} reports the full verifier-path
compute cost on the 3{,}168-video PhysicsIQ candidate pool
($N{=}16$ candidates $\times$ 198 scenarios). The body's
Fig.~\ref{fig:bon_compute} shows scoring time per video; this
table additionally reports parameters, VRAM, and total
wall-clock for each method.

\begin{table}[h]
\caption{\textbf{Scoring compute cost} on the 3{,}168-video
PhysicsIQ candidate pool. Scoring time and memory are for the
verifier path only, on top of the generator. All methods
benchmarked on a single H100 GPU.}
\label{tab:bon_compute_app}
\centering
\small
\setlength{\tabcolsep}{5pt}
\begin{tabular}{l l rrrr}
\toprule
Method & Verifier backbone & Params & VRAM & per video & total \\
\midrule
\multicolumn{6}{l}{\textit{Foundation-model baselines}} \\
VideoMAE(BoN)     & VideoMAE-large        & $0.3$B  & $1.4$\,GB  & $0.47$\,s & $25$\,min \\
Qwen2.5-VL(BoN)   & Qwen2.5-VL-7B-Inst.   & $7.0$B  & $16.6$\,GB & $0.28$\,s & $15$\,min \\
Qwen3-VL(BoN)     & Qwen3-VL-8B-Inst.     & $8.0$B  & $17.5$\,GB & $0.15$\,s & $8$\,min  \\
\midrule
\multicolumn{6}{l}{\textit{Trained world model}} \\
WMReward~\cite{yuan2026wmreward} & V-JEPA~2 ViT-giant & $1.1$B & $9.3$\,GB & $1.5$\,s & $77$\,min \\
\midrule
\multicolumn{6}{l}{\textit{\ours{} (frozen, no training)}} \\
\ours{} (1 backbone)   & DINOv3 ViT-L  & $0.3$B  & $1.2$\,GB  & $0.25$\,s & $13$\,min \\
\ours{} (4 backbones)  & Ensemble      & $0.7$B  & $2.0$\,GB  & $1.00$\,s & $53$\,min \\
\bottomrule
\end{tabular}
\end{table}

\newpage




\newpage
\newpage

\end{document}